\documentclass[11pt,table]{article}
\usepackage[final]{acl}

\usepackage{times}
\usepackage{latexsym}

\usepackage[T1]{fontenc}

\usepackage[utf8]{inputenc}

\usepackage{microtype}

\usepackage{inconsolata}

\usepackage{graphicx}
\usepackage{amsfonts}
\usepackage[most]{tcolorbox}
\usepackage{makecell}
\usepackage{xcolor}
\usepackage{subcaption} 

\title{Measuring Reasoning Utility in LLMs via Conditional Entropy Reduction}

\author{Xu Guo \\
  KTH Royal Institute of Technology, Sweden \\
  \texttt{xu008@e.ntu.edu.sg} \\}

\begin{document}
\maketitle
\begin{abstract}
Recent advancements in large language models (LLMs) often rely on generating intermediate reasoning steps to enhance accuracy. However, little work has examined how reasoning utility contributes to the final answer’s correctness. Due to the stochastic nature of autoregressive generation, generating more context does not guarantee increased confidence in the answer. If we could predict, during generation, whether a reasoning step will be useful, we could stop early or prune ineffective steps, avoiding distractions in the final decision.

We present an oracle study on MATH dataset, using Qwen2.5-32B and GPT-4o to generate reasoning chains, and then employing a separate model (Qwen3-8B) to quantify the utility of these chains for final accuracy. Specifically, we measure the model's uncertainty on the answer span $Y$ at each reasoning step using conditional entropy $\mathbb{E}_{v\in \mathcal{V}}
[-p(v|C)\log p(v|C)]$ with context $C$ expanding step by step. Our results show a clear pattern: conditional entropy that decreases over steps is strongly associated with correct answers, whereas flat or increasing entropy often results in wrong answers. We also corroborate that incorrect reasoning paths tend to be longer than correct ones, suggesting that longer reasoning does not necessarily yield better outcomes. These findings serve as a foundation to inspire future work on designing efficient reasoning pipelines that detect and avoid unproductive reasoning early.

\end{abstract}

\section{Introduction}

Recent progress in large language models (LLMs) has been driven largely by scaling model size, yielding unprecedented performance on benchmarks \cite{brown2020languagemodelsfewshotlearners,openai2024gpt4technicalreport,yang2025qwen3technicalreport}. However, scaling benefits show diminishing returns, and further gains from increasing parameters are costly. Meanwhile, such scaling has revealed emergent abilities that can push performance limits without additional training \cite{fast-BoN}. A notable finding is Chain-of-Thought (CoT) prompting \cite{wei2022chain}, which prompts the model to always generate intermediate thoughts before providing the answer. The use of CoT has been widely proven effective, inspiring research into advanced reasoning approaches \cite{yao2023tree,xu-etal-2025-softcot,xu2025softcottesttimescalingsoft}, agentic frameworks \cite{guo2024largelanguagemodelbased}, as well as models that are specially trained to think before answering \cite{openai2024openaio1card,deepseekai2025deepseekv3technicalreport}.

Yet, unrestricted generation of longer reasoning does not guarantee proportional accuracy gains. Techniques such as self-consistency \citep{wang2023selfconsistencyimproveschainthought} and best-of-N \citep{BoN} sampling exhibit diminishing returns once the number of sampled chains exceeds a certain threshold (e.g., >40), making them computationally inefficient. If the utility of reasoning for the final answer could be estimated during generation, unproductive reasoning could be halted early, reducing both distractions to the final answer and computational cost.

Prior work on measuring reasoning utility is scarce. Existing approaches fall into two categories: model-based, which rely on a verifier or pretrained Process Reward Model \cite{reward-model,fast-BoN}, and metric-based, which use statistical measures such as similarity or confidence. Similarity-based methods primarily target pruning redundancy, which does not directly link to accuracy \cite{path-consistency-SC,ST-BoN,internal-consistency}, as we corroborate in appendix \ref{app:sim}. Confidence-based metrics, such as perplexity and cross-entropy, are based on $p(y=y_{\mathrm{true}}|C)$, reflecting the model's confidence in the answer \cite{RPC,confidence-SC}. But our study in appendix~\ref{app:cross} finds limited evidence of its correlation with accuracy. In summary, there is a lack of systematic studies that examined how the model's uncertainty evolves as reasoning unfolds, and how this trajectory correlates with accuracy.

In this paper, we present an oracle study on the MATH \cite{hendrycks2021measuring} dataset across seven subjects. We use a reasoning LLM to generate CoT responses, segment them into steps, and compute conditional entropy on the answer span after each step. By grouping the responses into correct and incorrect sets, we plot the entropy trajectories and align them using cubic spline interpolation to reveal general trends across reasoning steps.

Our results on GPT-4o and Qwen2.5-32B reveal a strong correlation: when conditional entropy decreases over steps, the model is more likely to produce the correct answer; when it does not decrease, errors are more common. Also, incorrect reasoning paths are longer on average, indicating that more steps do not necessarily improve outcomes. These findings validate conditional entropy as a promising metric for reasoning utility, providing a foundation for future work on efficient reasoning pipelines that detect and avoid unproductive reasoning early.

\section{Preliminary}
We consider an autoregressive language model with vocabulary $\mathcal{V}$. 
Given a natural language prompt $s$, the tokenizer first maps it into a sequence of $N$ token IDs $\mathbf{x}_{1:N} = [x_1, \dots, x_N]$, where $x_i \in \mathcal{V}$.  
Each token ID is mapped to an embedding vector $\mathbf{e}_i \in \mathbb{R}^d$ through the embedding table $E \in \mathbb{R}^{|\mathcal{V}|\times d}$, producing the embedded input sequence
\[
X = [\mathbf{e}_1, \dots, \mathbf{e}_N] \in \mathbb{R}^{N\times d}.
\]

\noindent In autoregressive generation, the LLM predicts the next token distribution conditioned on a context $C$. Let $C_t$ denote the context at step $t$, which initially consists of $X$ and grows as new tokens are generated. For $t$-th token, the model outputs
\[
p_t(v) = \Pr(v \mid C_{<t}), \quad v \in \mathcal{V}.
\]
where $C_{<t}=[X; R_{<t}]$ and $R_{<t}$ denotes all previously generated tokens.  Generation stops when an end-of-sequence token is produced or a predefined maximum length is reached.

\noindent\textbf{Output structure.}
In our setting, the model’s complete response sequence is denoted as
\[
R=(z_1,z_2,\dots,z_L,y_1,y_2,\dots,y_M)
\]
where $Z=(z_1,z_2,\dots,z_L)$ is the reasoning chain, consisting of intermediate steps, explanations, or derivations. $Y=(y_1,y_2,\dots,y_M)$ is the final answer span.
This explicit decomposition allows us to analyze how the reasoning prefix $Z$ affects the model’s certainty about the final answer $Y$.

\section{Measuring the Utility of Reasoning}
Given that both the reasoning chain $Z$ and the final answer $Y$ are conditioned on the same input $X$, the joint distribution factorizes as: 
\[
P(Y,Z|X) = P(Z|X)\cdot P(Y|X,Z).
\]
Our key question is: \textit{``Given $X$ and a reasoning chain $Z$, how much additional information does $Z$ provide about the correct answer $Y$ beyond what is already contained in $X$?"} 

\noindent\textbf{Information-theoretic definition:} We formalize the utility of $Z$ via conditional mutual information:
\[
I(Y;Z|X) = H(Y \mid X) - H(Y \mid X, Z).
\]
where $H(\cdot)$ denotes entropy.
A large reduction in conditional entropy $\Delta H$ indicates that the reasoning chain $Z$ significantly narrows down the correct answer space, while $\Delta H \approx 0$ means $Z$ adds little useful information beyond the question itself.

\noindent\textbf{Practical estimation via conditional entropy:} Directly computing $H(Y|X)$ and $H(Y|X,Z)$ is intractable for LLMs. We therefore estimate them using teacher forcing: For each token position $t$ in a target answer sequence $Y = (y_1, \ldots, y_{|Y|})$ and for a given context $C = X + Z_{\leq k}$ at $k$-th reasoning step, we feed the LLM $[C; y_{<t}]$ to obtain
\[
p_t(v) = \Pr(v \mid C, y_{<t}), \quad v \in \mathcal{V}.
\]
The token-level entropy at position $t$ is:
\[
H_t = - \sum_{v \in \mathcal{V}} p_t(v) \log p_t(v).
\]
Since the ground-truth $Y$ is a sequence, we define the \textbf{sequence-level conditional entropy} as the average token entropy: 
\begin{equation}\label{eq:answer-entropy}
H(Y \mid C) = \frac{1}{|Y|} \sum_{t=1}^{|Y|} H_t
= \mathbb{E}_{t \in {1,\ldots, |Y|}} \left[ H_t \right].
\end{equation}
This conditional entropy directly reflects the model’s average uncertainty about $Y$ given $C$. Notably, this measurement relies solely on the model’s own output probabilities and requires no additional reward model or external supervision. 

\noindent\textbf{Entropy trajectory during reasoning:} 
To study how reasoning affects answer certainty, we decompose $Z$ into $K$ steps and compute $H(Y \mid X, Z_{\le k})$ for each $k = 0, 1, \ldots, K$, where $Z_{\le k}$ denotes the first $k$ reasoning steps. This yields an \textit{entropy trajectory} that reflects how uncertainty changes as reasoning unfolds. The information gain at $k$-th reasoning step is therefore estimated by $H(Y \mid X, Z_{\le k-1}) - H(Y \mid X, Z_{\le k})$.

\begin{table*}[!h]
    \centering
     \small
    \begin{tabular}{l|l|l|l|l|l|l|l}\hline
     & \makecell{\textbf{Counting \&} \\ \textbf{Probability}}  & \makecell{\textbf{Number} \\ \textbf{Theory}} & \textbf{Prealgebra} & \textbf{Algebra} & \makecell{\textbf{Intermediate} \\ \textbf{Algebra}} & \textbf{Precalculus} & \textbf{Geometry} \\\hline
     No. of problems & 469 & 540  & 864 & 1185 & 903 & 546& 479\\\hline
     \makecell{Avg. No. of tokens \\ in human solutions} & 466 & 472 & 357 & 370 & 660 & 780& 726 \\\hline
     \makecell{Avg. steps ($\overline{K}_{\textrm{gt}}$) of \\ human solutions} & 5 & 5 & 5 & 4 & 7 & 7 & 8 \\\hline\hline
      \makecell{Avg. No. of tokens \\ in LLM solutions} & 1728 & 1565 & 1281 & 1306 & 1860 & 1930& 1895 \\\hline
      \makecell{Avg. steps ($\overline{K}_{\textrm{llm}}$) of \\ LLM solutions} & 9 & 10  & 9 & 10 & 11 & 10 & 10 \\\hline
      \rowcolor{blue!10}
      LLM Accuracy & 0.81 &  0.84 & 0.92 & 0.95 & 0.63 & 0.63 & 0.67 \\\hline
    \end{tabular}
    \caption{Statistics of the MATH dataset and a comparison between human and LLM (Qwen2.5-32B) solutions. LLM solutions contain substantially more tokens than human solutions. We report LLM accuracy against human solutions as the ground truth, and observe that lower-accuracy categories tend to involve longer reasoning.}
    \label{tab:data_stats}
\end{table*}

\section{Experiment}
\paragraph{Dataset.} We use the MATH dataset \cite{hendrycks2021measuring}, which spans seven math domains, across five difficulty levels. Each problem is provided with a human step-by-step solution. Summary statistics are shown in Table \ref{tab:data_stats}.

\paragraph{Oracle study setting.}
For data generation and inspection, we employ different LLMs. Specifically, we use GPT-4o and Qwen2.5-32B to generate model-produced solutions for each problem and compare them with the human references. To measure uncertainty, we use Qwen3-8B as an \emph{inspector} that computes answer-span conditional entropy under teacher forcing. We simulate autoregressive generation, and at each token position $t$ in the answer span, we use the last-layer last-token hidden state to obtain the logits for computing the entropy.

\begin{figure*}[htbp]
  \centering
  \begin{subfigure}[t]{0.98\textwidth}
    \centering
    \includegraphics[width=\textwidth]{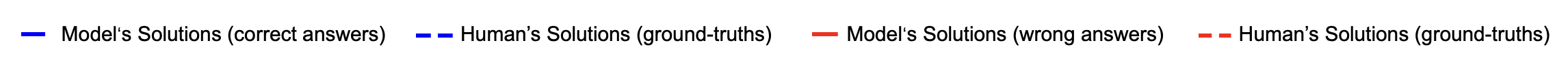}
  \end{subfigure}
  \hfill
  
  \begin{subfigure}[t]{0.47\textwidth}
    \centering
    \includegraphics[width=\textwidth]{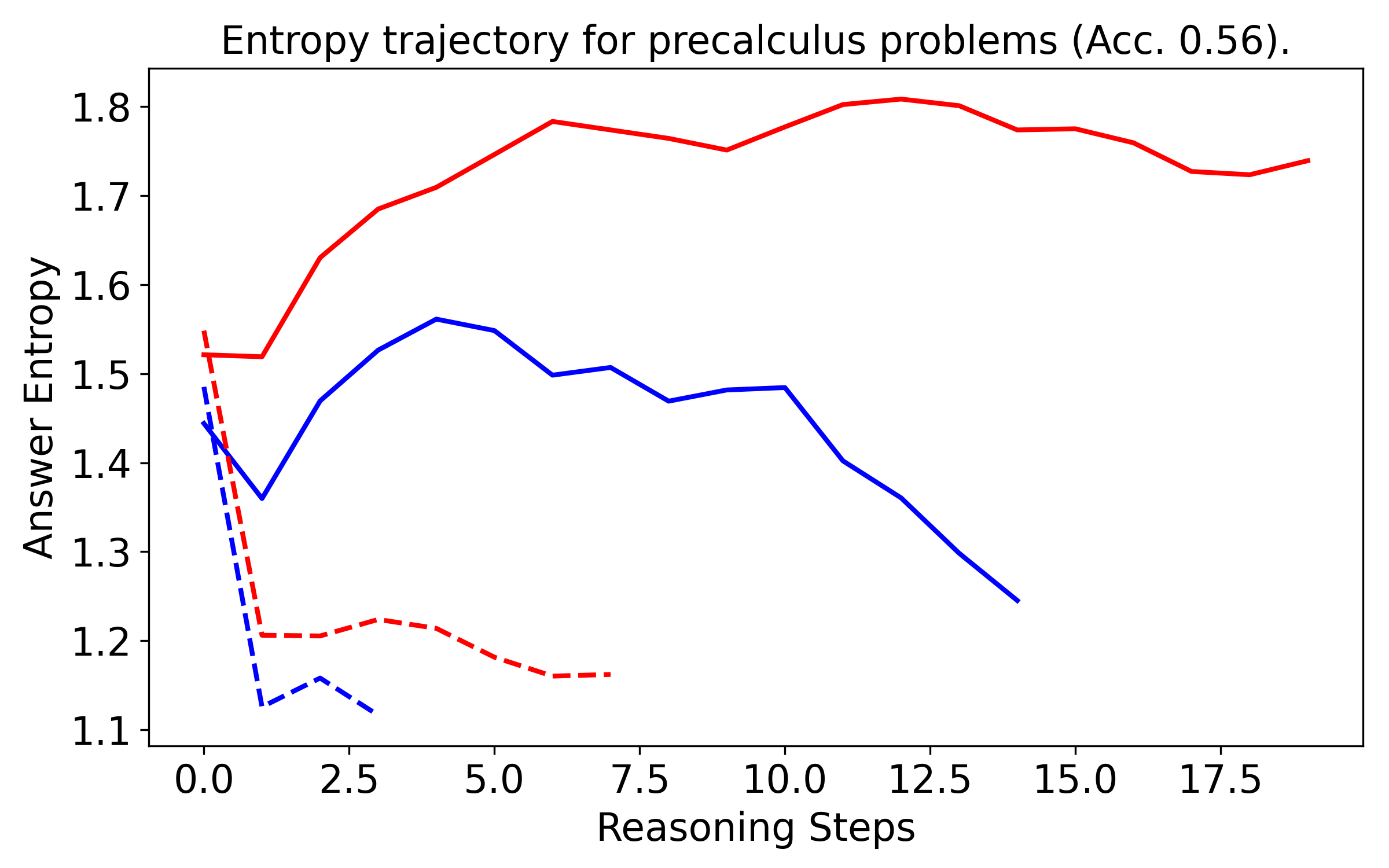}
    \caption{GPT-4o vs. Human Reasoning on Precalculus. }
    \label{fig:subfig1}
  \end{subfigure}
  \hfill
  \begin{subfigure}[t]{0.47\textwidth}
    \centering
    \includegraphics[width=\textwidth]{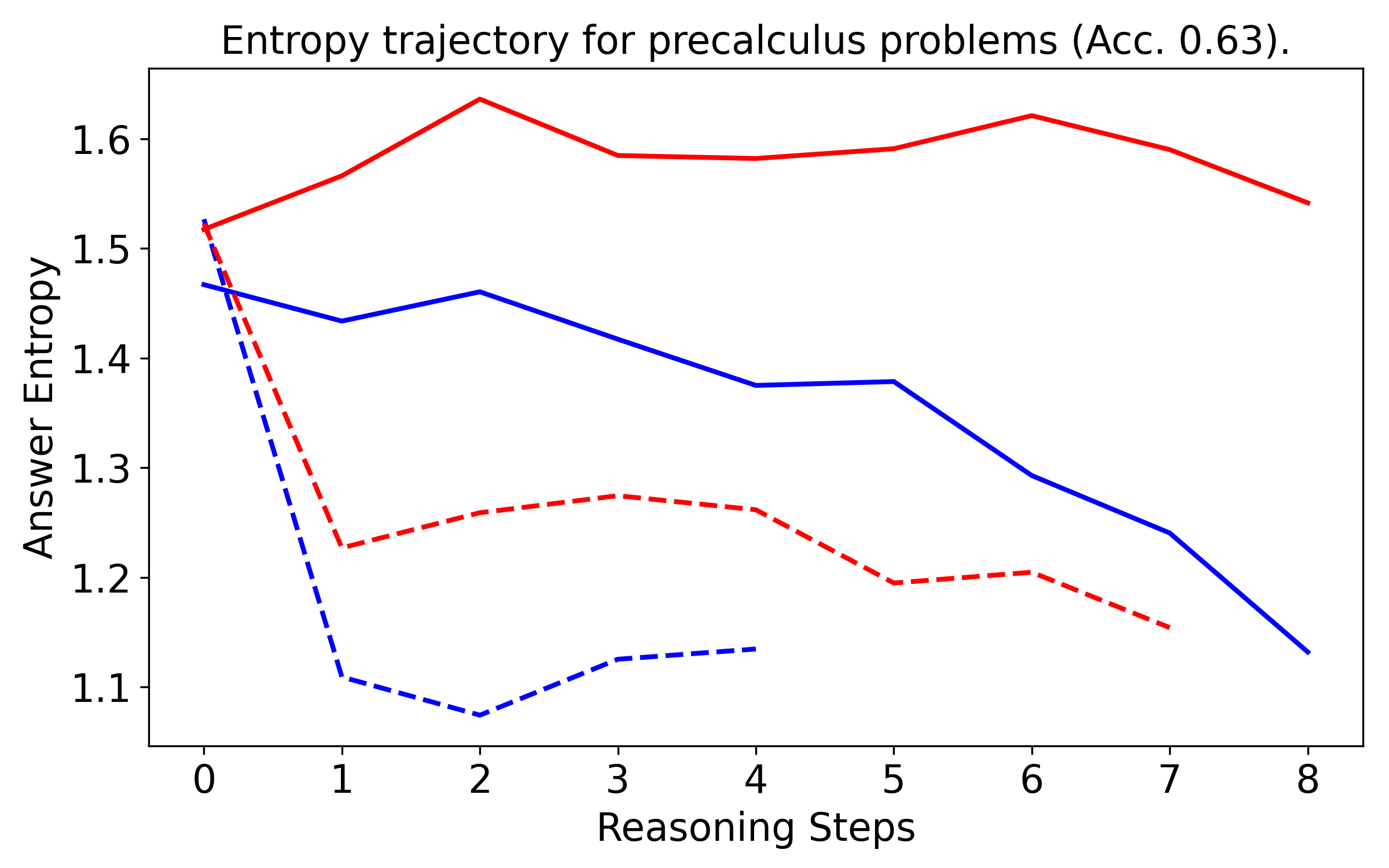}
    \caption{Qwen2.5-32B vs. Human Reasoning on Precalculus. }
    \label{fig:subfig2}
  \end{subfigure}
  \hfill
  \begin{subfigure}[t]{0.47\textwidth}
    \centering
    \includegraphics[width=\textwidth]{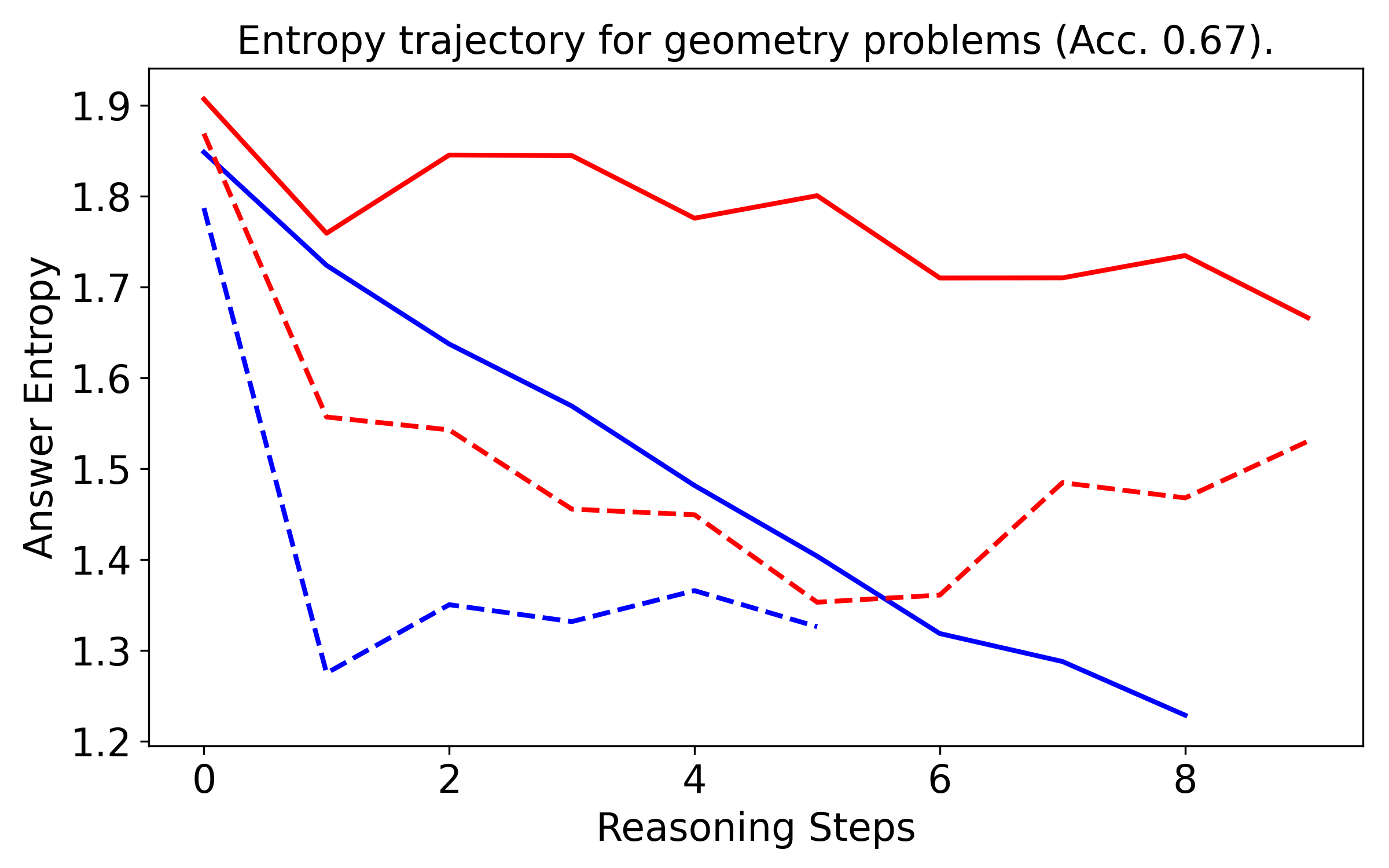}
    \caption{Qwen2.5-32B vs. Human Reasoning on Geometry.}
    \label{fig:subfig3}
  \end{subfigure}
  \hfill
  \begin{subfigure}[t]{0.47\textwidth}
    \centering
    \includegraphics[width=\textwidth]{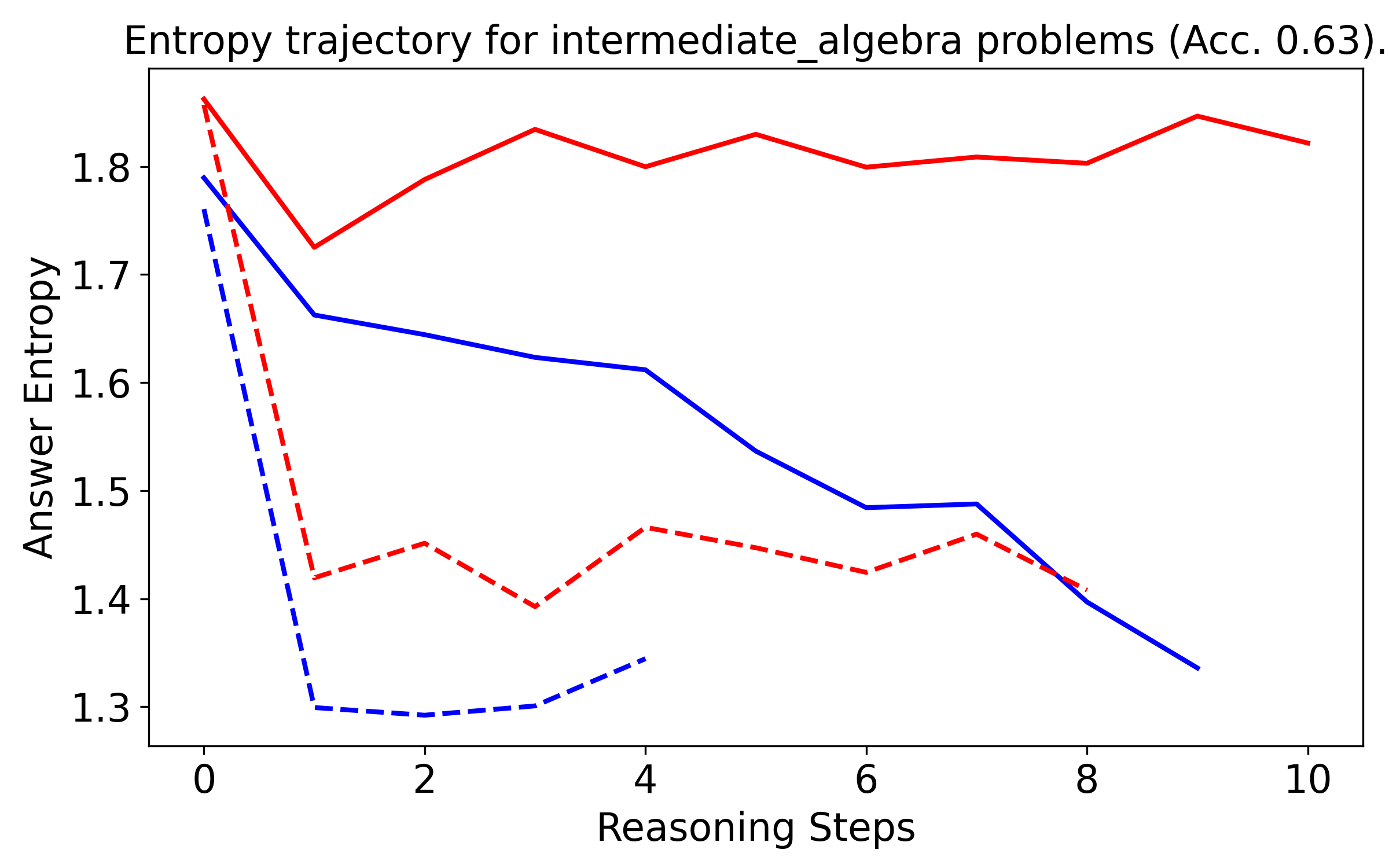}
    \caption{Qwen2.5-32B vs. Human on Intermediate Algebra.}
    \label{fig:subfig4}
  \end{subfigure}
  \hfill
  \begin{subfigure}[t]{0.47\textwidth}
    \centering
    \includegraphics[width=\textwidth]{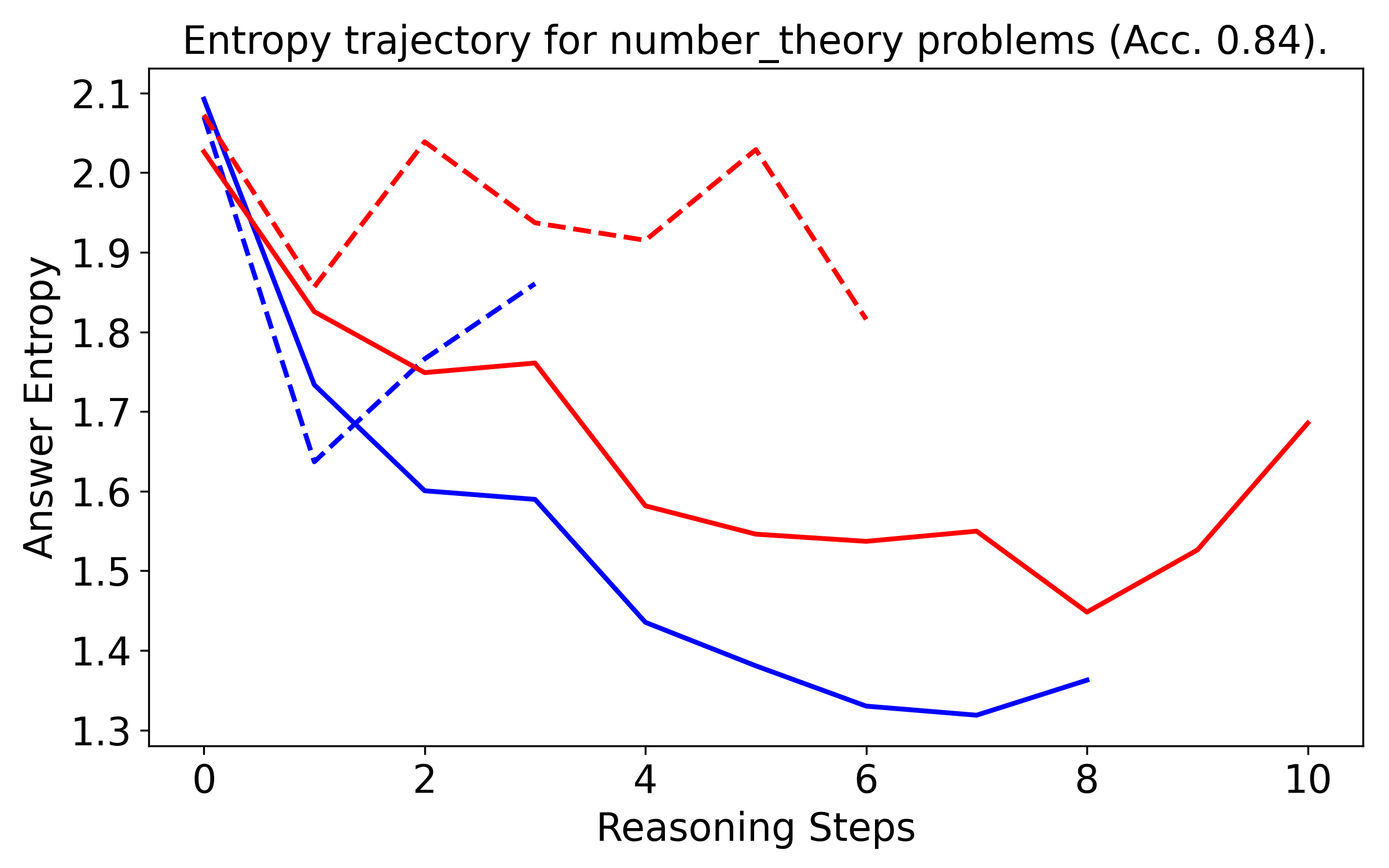}
    \caption{Qwen2.5-32B vs. Human Reasoning on Number Theory.}
    \label{fig:subfig5}
  \end{subfigure}
  \hfill
  \begin{subfigure}[t]{0.47\textwidth}
    \centering
    \includegraphics[width=\textwidth]{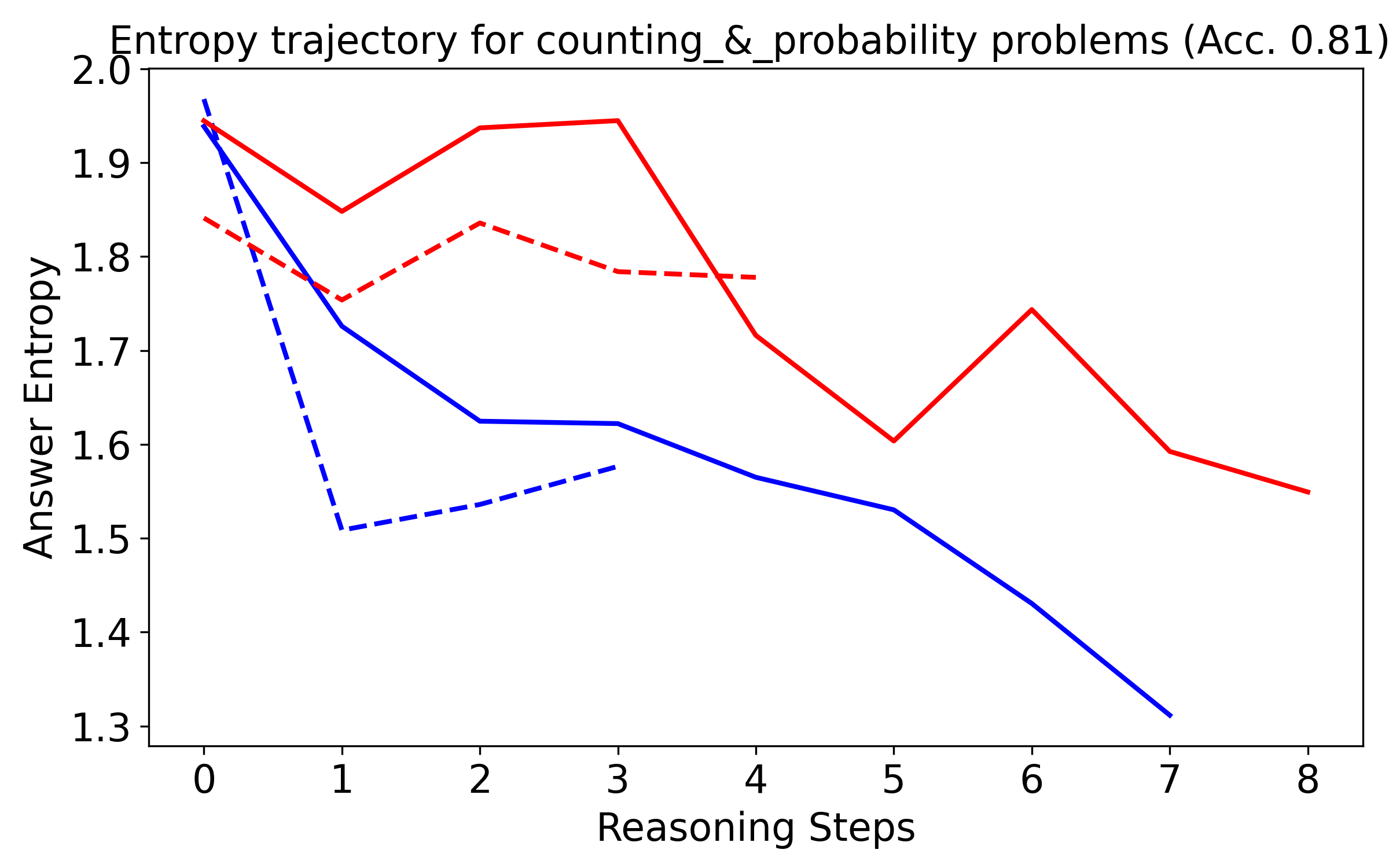}
    \caption{Qwen2.5-32B vs. Human on Counting \& Probability.}
    \label{fig:subfig6}
  \end{subfigure}
  \hfill
  \begin{subfigure}[t]{0.47\textwidth}
    \centering
    \includegraphics[width=\textwidth]{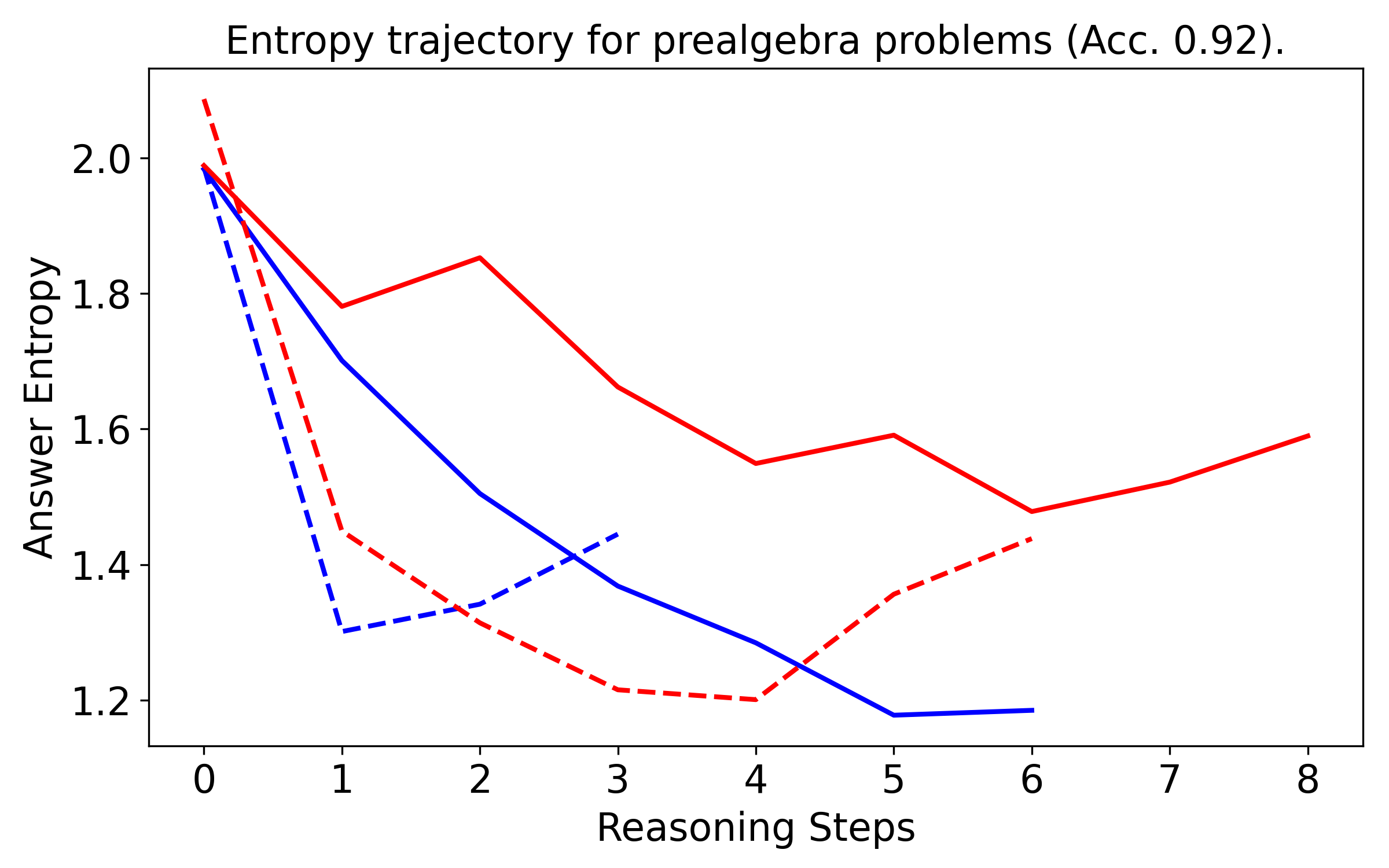}
    \caption{Qwen2.5-32B vs. Human Reasoning on Prealgebra.}
    \label{fig:subfig7}
  \end{subfigure}
  \hfill
  \begin{subfigure}[t]{0.47\textwidth}
    \centering
    \includegraphics[width=\textwidth]{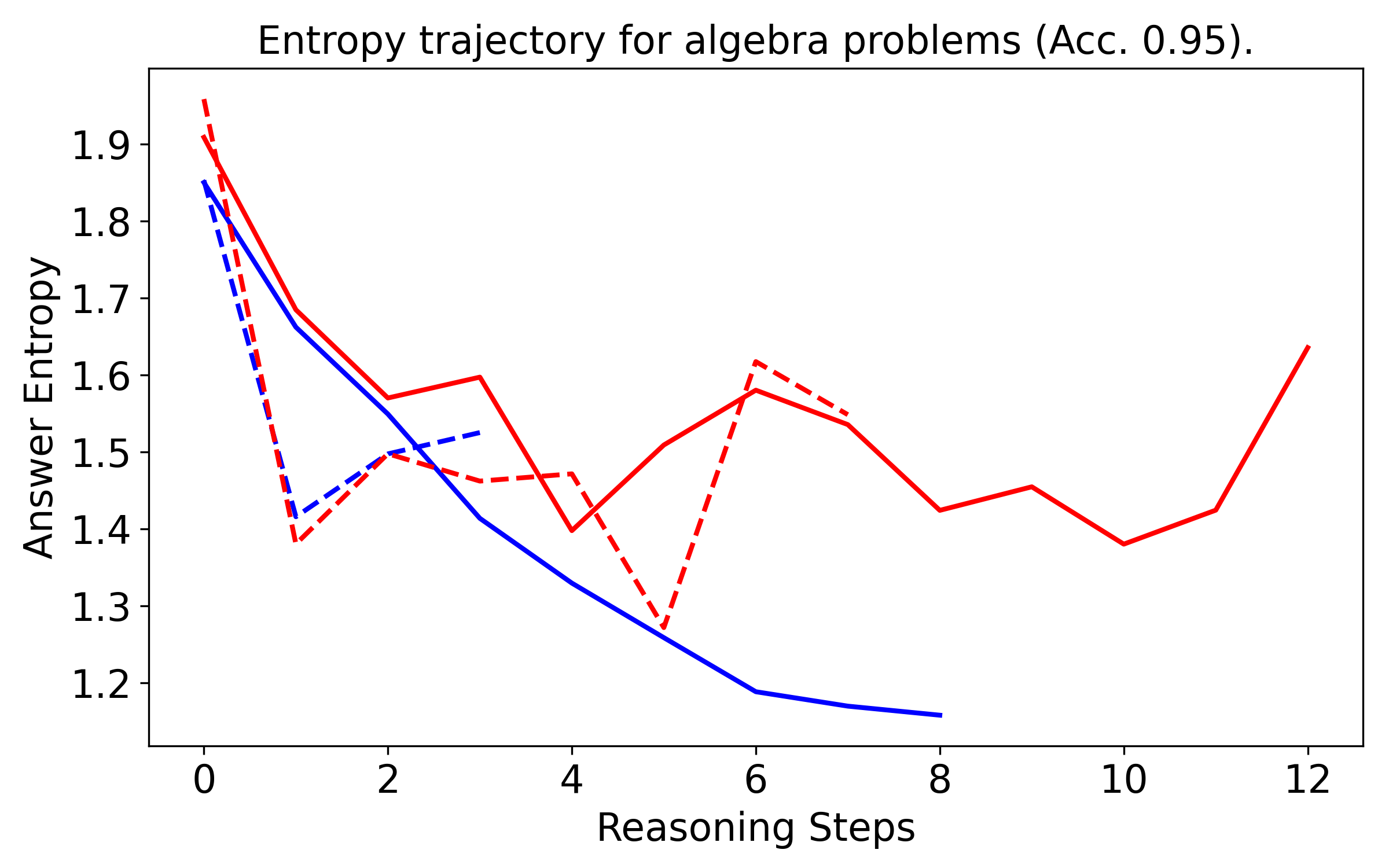}
    \caption{Qwen2.5-32B vs. Human Reasoning on Algebra.}
    \label{fig:subfig8}
  \end{subfigure}
  \caption{Entropy variation across reasoning steps. Sub-fig. (a-b) compare two LLMs on the same set of math problems; Sub-fig. (c-h) show the same LLM across problem types. \textbf{Every single curve is an average entropy trajectory over a set of problems, aligned on reasoning steps using cubic-spline interpolation.} Solid blue curves \textcolor{blue}{---} end with correct LLM answers. Solid red curves \textcolor{red}{---} end with incorrect LLM answers. Dashed curves of the same color (\textcolor{blue}{-- --} and \textcolor{red}{-- --}) are their corresponding human (ground-truth) solutions for the same problem subsets. \textbf{Main observations}: correct trajectories exhibit a steeper, earlier entropy decrease than incorrect ones; in high-accuracy categories (Sub-fig. e-h), correct trajectories remain below incorrect ones, whereas in lower-accuracy categories (Sub-fig. a-d), incorrect trajectories generally do not decrease. (More results for GPT-4o are presented in Figure \ref{fig:overall-gpt4o})}
  \label{fig:overall}
\end{figure*}

\paragraph{Data Processing.}
Every reasoning chain can have a different number of steps, making direct averaging on the same problem set infeasible.
For each domain, we align entropy trajectories by resampling each chain to a common target steps ($\overline{K}_{\text{gt}}$ for human and $\overline{K}_{\text{llm}}$ for LLM). We use cubic splines for resampling (switch to linear interpolation when steps $\le 3$) to place all curves on a shared horizontal axis.
This up‑/down‑sampling preserves the overall shape of each trajectory while giving all curves the same number of points.  
After alignment, we average entropy at each normalised step to obtain domain-level trends, as shown in Figure~\ref{fig:overall}.

\section{Results Discussion and Opportunities}\label{sec:results}
This oracle study presents the following insights.
\noindent\textbf{Reasoning paths that lead to correct answers consistently achieve information gain over steps.}
Across seven math categories, the answer-span entropy for correct LLM chains (\textcolor{blue}{—}) consistently decreases, falls faster, and remains below that of incorrect chains (\textcolor{red}{—}). In contrast, incorrect chains show no consistent decrease: on lower-accuracy categories (Fig.\ref{fig:subfig1}-\ref{fig:subfig4}) entropy is typically flat or even rises, while on higher-accuracy categories (Fig.\ref{fig:subfig5}-\ref{fig:subfig8}) it may decrease yet stays above the correct trajectories. These patterns suggest a simple selection heuristic: prune chains whose entropy does not decrease; if all decrease, rank by the magnitude of the negative slope and keep the top-$k$.

\noindent\textbf{Reasoning paths that lead to wrong answers are typically longer.}
Across the same problem sets, LLM-generated chains (solid) are generally longer than human chains (dashed), with GPT-4o (Fig.\ref{fig:subfig1}) exhibiting longer chains than Qwen2.5-32B (Fig.\ref{fig:subfig2}). Incorrect chains (red) are significantly longer than correct ones (blue). These findings suggest that LLMs often fail on problems that require more steps, and that tackling hard problems requires more advanced reasoning strategies instead of simply scaling the number of steps.

\noindent\textbf{Information gains diminish faster for human reasoning than for LLM reasoning.}
In human solutions, entropy drops sharply in the early steps and then plateaus; by contrast, correct LLM trajectories (solid blue) decrease entropy more gradually and uniformly across steps. This pattern suggests a strategic difference: humans quickly narrow the answer space and then unravel the intermediate steps, whereas LLMs show weaker early narrowing and progress step by step until the answer becomes clear, as illustrated by an example in Appendix~\ref{app:example}. This suggests that improving LLMs’ ability to narrow the answer space early, instead of planning step by step, could facilitate efficient reasoning.


\section*{Limitations}
While our study supports that entropy trajectories provide a model-internal, training-free signal to assess reasoning utility, several limitations remain: 
\begin{itemize}

    \item Our analysis relies on access to ground-truth answers, which enables controlled comparison between human and model trajectories. In real deployment, however, ground truth is unavailable, so entropy-based signals must be interpreted without oracle guidance. This limits the direct applicability of our findings, although they still provide useful guidance for pruning parallel scaling strategies (e.g., self-consistency, best-of-$N$) where answer consensus is the main signal.

    \item Experiments are conducted on the MATH dataset. While math problems are a natural testbed for reasoning, we need more dataset evaluations to support whether the observed entropy dynamics generalize to other domains, such as commonsense reasoning.
    \item We use Qwen3-8B as the inspector model. We need more evaluations on different architectures or sizes to confirm the generalizability of the entropy dynamics.
    
\end{itemize}


\bibliography{main}

\begin{thebibliography}{20}
\providecommand{\natexlab}[1]{#1}

\bibitem[{Brown et~al.(2020)Brown, Mann, Ryder, Subbiah, Kaplan, Dhariwal, Neelakantan, Shyam, Sastry, Askell, Agarwal, Herbert-Voss, Krueger, Henighan, Child, Ramesh, Ziegler, Wu, Winter, Hesse, Chen, Sigler, Litwin, Gray, Chess, Clark, Berner, McCandlish, Radford, Sutskever, and Amodei}]{brown2020languagemodelsfewshotlearners}
Tom~B. Brown, Benjamin Mann, Nick Ryder, Melanie Subbiah, Jared Kaplan, Prafulla Dhariwal, Arvind Neelakantan, Pranav Shyam, Girish Sastry, Amanda Askell, Sandhini Agarwal, Ariel Herbert-Voss, Gretchen Krueger, Tom Henighan, Rewon Child, Aditya Ramesh, Daniel~M. Ziegler, Jeffrey Wu, Clemens Winter, and 12 others. 2020.
\newblock \href {https://arxiv.org/abs/2005.14165} {Language models are few-shot learners}.
\newblock \emph{Preprint}, arXiv:2005.14165.

\bibitem[{Cobbe et~al.(2021)Cobbe, Kosaraju, Bavarian, Chen, Jun, Kaiser, Plappert, Tworek, Hilton, Nakano, Hesse, and Schulman}]{reward-model}
Karl Cobbe, Vineet Kosaraju, Mohammad Bavarian, Mark Chen, Heewoo Jun, Lukasz Kaiser, Matthias Plappert, Jerry Tworek, Jacob Hilton, Reiichiro Nakano, Christopher Hesse, and John Schulman. 2021.
\newblock \href {https://arxiv.org/abs/2110.14168} {Training verifiers to solve math word problems}.
\newblock \emph{Preprint}, arXiv:2110.14168.

\bibitem[{DeepSeek-AI et~al.(2025)DeepSeek-AI, Liu, Feng, Xue, Wang, Wu, Lu, Zhao, Deng, Zhang, Ruan, Dai, Guo, Yang, Chen, Ji, Li, Lin, Dai, Luo, Hao, Chen, Li, Zhang, Bao, Xu, Wang, Zhang, Ding, Xin, Gao, Li, Qu, Cai, Liang, Guo, Ni, Li, Wang, Chen, Chen, Yuan, Qiu, Li, Song, Dong, Hu, Gao, Guan, Huang, Yu, Wang, Zhang, Xu, Xia, Zhao, Wang, Zhang, Li, Wang, Zhang, Zhang, Tang, Li, Tian, Huang, Wang, Zhang, Wang, Zhu, Chen, Du, Chen, Jin, Ge, Zhang, Pan, Wang, Xu, Zhang, Chen, Li, Lu, Zhou, Chen, Wu, Ye, Ye, Ma, Wang, Zhou, Yu, Zhou, Pan, Wang, Yun, Pei, Sun, Xiao, Zeng, Zhao, An, Liu, Liang, Gao, Yu, Zhang, Li, Jin, Wang, Bi, Liu, Wang, Shen, Chen, Zhang, Chen, Nie, Sun, Wang, Cheng, Liu, Xie, Liu, Yu, Song, Shan, Zhou, Yang, Li, Su, Lin, Li, Wang, Wei, Zhu, Zhang, Xu, Xu, Huang, Li, Zhao, Sun, Li, Wang, Yu, Zheng, Zhang, Shi, Xiong, He, Tang, Piao, Wang, Tan, Ma, Liu, Guo, Wu, Ou, Zhu, Wang, Gong, Zou, He, Zha, Xiong, Ma, Yan, Luo, You, Liu, Zhou, Wu, Ren, Ren, Sha, Fu, Xu, Huang, Zhang, Xie, Zhang, Hao,
  Gou, Ma, Yan, Shao, Xu, Wu, Zhang, Li, Gu, Zhu, Liu, Li, Xie, Song, Gao, and Pan}]{deepseekai2025deepseekv3technicalreport}
DeepSeek-AI, Aixin Liu, Bei Feng, Bing Xue, Bingxuan Wang, Bochao Wu, Chengda Lu, Chenggang Zhao, Chengqi Deng, Chenyu Zhang, Chong Ruan, Damai Dai, Daya Guo, Dejian Yang, Deli Chen, Dongjie Ji, Erhang Li, Fangyun Lin, Fucong Dai, and 181 others. 2025.
\newblock \href {https://arxiv.org/abs/2412.19437} {Deepseek-v3 technical report}.
\newblock \emph{Preprint}, arXiv:2412.19437.

\bibitem[{Guo et~al.(2024)Guo, Chen, Wang, Chang, Pei, Chawla, Wiest, and Zhang}]{guo2024largelanguagemodelbased}
Taicheng Guo, Xiuying Chen, Yaqi Wang, Ruidi Chang, Shichao Pei, Nitesh~V. Chawla, Olaf Wiest, and Xiangliang Zhang. 2024.
\newblock \href {https://arxiv.org/abs/2402.01680} {Large language model based multi-agents: A survey of progress and challenges}.
\newblock \emph{Preprint}, arXiv:2402.01680.

\bibitem[{Hendrycks et~al.(2021)Hendrycks, Burns, Kadavath, Arora, Basart, Tang, Song, and Steinhardt}]{hendrycks2021measuring}
Dan Hendrycks, Collin Burns, Saurav Kadavath, Akul Arora, Steven Basart, Eric Tang, Dawn Song, and Jacob Steinhardt. 2021.
\newblock \href {https://openreview.net/forum?id=7Bywt2mQsCe} {Measuring mathematical problem solving with the {MATH} dataset}.
\newblock In \emph{Thirty-fifth Conference on Neural Information Processing Systems Datasets and Benchmarks Track (Round 2)}.

\bibitem[{Lightman et~al.(2024)Lightman, Kosaraju, Burda, Edwards, Baker, Lee, Leike, Schulman, Sutskever, and Cobbe}]{BoN}
Hunter Lightman, Vineet Kosaraju, Yuri Burda, Harrison Edwards, Bowen Baker, Teddy Lee, Jan Leike, John Schulman, Ilya Sutskever, and Karl Cobbe. 2024.
\newblock \href {https://openreview.net/forum?id=v8L0pN6EOi} {Let's verify step by step}.
\newblock In \emph{The Twelfth International Conference on Learning Representations}.

\bibitem[{OpenAI et~al.(2024{\natexlab{a}})OpenAI, :, Jaech, Kalai, Lerer, Richardson, El-Kishky, Low, Helyar, Madry, Beutel, Carney, Iftimie, Karpenko, Passos, Neitz, Prokofiev, Wei, Tam, Bennett, Kumar, Saraiva, Vallone, Duberstein, Kondrich, Mishchenko, Applebaum, Jiang, Nair, Zoph, Ghorbani, Rossen, Sokolowsky, Barak, McGrew, Minaiev, Hao, Baker, Houghton, McKinzie, Eastman, Lugaresi, Bassin, Hudson, Li, de~Bourcy, Voss, Shen, Zhang, Koch, Orsinger, Hesse, Fischer, Chan, Roberts, Kappler, Levy, Selsam, Dohan, Farhi, Mely, Robinson, Tsipras, Li, Oprica, Freeman, Zhang, Wong, Proehl, Cheung, Mitchell, Wallace, Ritter, Mays, Wang, Such, Raso, Leoni, Tsimpourlas, Song, von Lohmann, Sulit, Salmon, Parascandolo, Chabot, Zhao, Brockman, Leclerc, Salman, Bao, Sheng, Andrin, Bagherinezhad, Ren, Lightman, Chung, Kivlichan, O'Connell, Osband, Gilaberte, Akkaya, Kostrikov, Sutskever, Kofman, Pachocki, Lennon, Wei, Harb, Twore, Feng, Yu, Weng, Tang, Yu, Candela, Palermo, Parish, Heidecke, Hallman, Rizzo, Gordon,
  Uesato, Ward, Huizinga, Wang, Chen, Xiao, Singhal, Nguyen, Cobbe, Shi, Wood, Rimbach, Gu-Lemberg, Liu, Lu, Stone, Yu, Ahmad, Yang, Liu, Maksin, Ho, Fedus, Weng, Li, McCallum, Held, Kuhn, Kondraciuk, Kaiser, Metz, Boyd, Trebacz, Joglekar, Chen, Tintor, Meyer, Jones, Kaufer, Schwarzer, Shah, Yatbaz, Guan, Xu, Yan, Glaese, Chen, Lampe, Malek, Wang, Fradin, McClay, Pavlov, Wang, Wang, Murati, Bavarian, Rohaninejad, McAleese, Chowdhury, Chowdhury, Ryder, Tezak, Brown, Nachum, Boiko, Murk, Watkins, Chao, Ashbourne, Izmailov, Zhokhov, Dias, Arora, Lin, Lopes, Gaon, Miyara, Leike, Hwang, Garg, Brown, James, Shu, Cheu, Greene, Jain, Altman, Toizer, Toyer, Miserendino, Agarwal, Hernandez, Baker, McKinney, Yan, Zhao, Hu, Santurkar, Chaudhuri, Zhang, Fu, Papay, Lin, Balaji, Sanjeev, Sidor, Broda, Clark, Wang, Gordon, Sanders, Patwardhan, Sottiaux, Degry, Dimson, Zheng, Garipov, Stasi, Bansal, Creech, Peterson, Eloundou, Qi, Kosaraju, Monaco, Pong, Fomenko, Zheng, Zhou, McCabe, Zaremba, Dubois, Lu, Chen, Cha, Bai, He,
  Zhang, Wang, Shao, and Li}]{openai2024openaio1card}
OpenAI, :, Aaron Jaech, Adam Kalai, Adam Lerer, Adam Richardson, Ahmed El-Kishky, Aiden Low, Alec Helyar, Aleksander Madry, Alex Beutel, Alex Carney, Alex Iftimie, Alex Karpenko, Alex~Tachard Passos, Alexander Neitz, Alexander Prokofiev, Alexander Wei, Allison Tam, and 244 others. 2024{\natexlab{a}}.
\newblock \href {https://arxiv.org/abs/2412.16720} {Openai o1 system card}.
\newblock \emph{Preprint}, arXiv:2412.16720.

\bibitem[{OpenAI et~al.(2024{\natexlab{b}})OpenAI, Achiam, Adler, Agarwal, Ahmad, Akkaya, Aleman, Almeida, Altenschmidt, Altman, Anadkat, Avila, Babuschkin, Balaji, Balcom, Baltescu, Bao, Bavarian, Belgum, Bello, Berdine, Bernadett-Shapiro, Berner, Bogdonoff, Boiko, Boyd, Brakman, Brockman, Brooks, Brundage, Button, Cai, Campbell, Cann, Carey, Carlson, Carmichael, Chan, Chang, Chantzis, Chen, Chen, Chen, Chen, Chen, Chess, Cho, Chu, Chung, Cummings, Currier, Dai, Decareaux, Degry, Deutsch, Deville, Dhar, Dohan, Dowling, Dunning, Ecoffet, Eleti, Eloundou, Farhi, Fedus, Felix, Fishman, Forte, Fulford, Gao, Georges, Gibson, Goel, Gogineni, Goh, Gontijo-Lopes, Gordon, Grafstein, Gray, Greene, Gross, Gu, Guo, Hallacy, Han, Harris, He, Heaton, Heidecke, Hesse, Hickey, Hickey, Hoeschele, Houghton, Hsu, Hu, Hu, Huizinga, Jain, Jain, Jang, Jiang, Jiang, Jin, Jin, Jomoto, Jonn, Jun, Kaftan, Łukasz Kaiser, Kamali, Kanitscheider, Keskar, Khan, Kilpatrick, Kim, Kim, Kim, Kirchner, Kiros, Knight, Kokotajlo, Łukasz
  Kondraciuk, Kondrich, Konstantinidis, Kosic, Krueger, Kuo, Lampe, Lan, Lee, Leike, Leung, Levy, Li, Lim, Lin, Lin, Litwin, Lopez, Lowe, Lue, Makanju, Malfacini, Manning, Markov, Markovski, Martin, Mayer, Mayne, McGrew, McKinney, McLeavey, McMillan, McNeil, Medina, Mehta, Menick, Metz, Mishchenko, Mishkin, Monaco, Morikawa, Mossing, Mu, Murati, Murk, Mély, Nair, Nakano, Nayak, Neelakantan, Ngo, Noh, Ouyang, O'Keefe, Pachocki, Paino, Palermo, Pantuliano, Parascandolo, Parish, Parparita, Passos, Pavlov, Peng, Perelman, de~Avila Belbute~Peres, Petrov, de~Oliveira~Pinto, Michael, Pokorny, Pokrass, Pong, Powell, Power, Power, Proehl, Puri, Radford, Rae, Ramesh, Raymond, Real, Rimbach, Ross, Rotsted, Roussez, Ryder, Saltarelli, Sanders, Santurkar, Sastry, Schmidt, Schnurr, Schulman, Selsam, Sheppard, Sherbakov, Shieh, Shoker, Shyam, Sidor, Sigler, Simens, Sitkin, Slama, Sohl, Sokolowsky, Song, Staudacher, Such, Summers, Sutskever, Tang, Tezak, Thompson, Tillet, Tootoonchian, Tseng, Tuggle, Turley, Tworek, Uribe,
  Vallone, Vijayvergiya, Voss, Wainwright, Wang, Wang, Wang, Ward, Wei, Weinmann, Welihinda, Welinder, Weng, Weng, Wiethoff, Willner, Winter, Wolrich, Wong, Workman, Wu, Wu, Wu, Xiao, Xu, Yoo, Yu, Yuan, Zaremba, Zellers, Zhang, Zhang, Zhao, Zheng, Zhuang, Zhuk, and Zoph}]{openai2024gpt4technicalreport}
OpenAI, Josh Achiam, Steven Adler, Sandhini Agarwal, Lama Ahmad, Ilge Akkaya, Florencia~Leoni Aleman, Diogo Almeida, Janko Altenschmidt, Sam Altman, Shyamal Anadkat, Red Avila, Igor Babuschkin, Suchir Balaji, Valerie Balcom, Paul Baltescu, Haiming Bao, Mohammad Bavarian, Jeff Belgum, and 262 others. 2024{\natexlab{b}}.
\newblock \href {https://arxiv.org/abs/2303.08774} {Gpt-4 technical report}.
\newblock \emph{Preprint}, arXiv:2303.08774.

\bibitem[{Sun et~al.(2024)Sun, Haider, Zhang, Yang, Qiu, Yin, Wang, Bartlett, and Zanette}]{fast-BoN}
Hanshi Sun, Momin Haider, Ruiqi Zhang, Huitao Yang, Jiahao Qiu, Ming Yin, Mengdi Wang, Peter Bartlett, and Andrea Zanette. 2024.
\newblock \href {https://openreview.net/forum?id=348hfcprUs} {Fast best-of-n decoding via speculative rejection}.
\newblock In \emph{The Thirty-eighth Annual Conference on Neural Information Processing Systems}.

\bibitem[{Taubenfeld et~al.(2025)Taubenfeld, Sheffer, Ofek, Feder, Goldstein, Gekhman, and Yona}]{confidence-SC}
Amir Taubenfeld, Tom Sheffer, Eran Ofek, Amir Feder, Ariel Goldstein, Zorik Gekhman, and Gal Yona. 2025.
\newblock \href {https://arxiv.org/abs/2502.06233} {Confidence improves self-consistency in llms}.
\newblock \emph{Preprint}, arXiv:2502.06233.

\bibitem[{Wang et~al.(2023)Wang, Wei, Schuurmans, Le, Chi, Narang, Chowdhery, and Zhou}]{wang2023selfconsistencyimproveschainthought}
Xuezhi Wang, Jason Wei, Dale Schuurmans, Quoc Le, Ed~Chi, Sharan Narang, Aakanksha Chowdhery, and Denny Zhou. 2023.
\newblock \href {https://arxiv.org/abs/2203.11171} {Self-consistency improves chain of thought reasoning in language models}.
\newblock \emph{Preprint}, arXiv:2203.11171.

\bibitem[{Wang et~al.(2025)Wang, Zhang, Huang, Yang, Zhang, Huang, and Wang}]{ST-BoN}
Yiming Wang, Pei Zhang, Siyuan Huang, Baosong Yang, Zhuosheng Zhang, Fei Huang, and Rui Wang. 2025.
\newblock \href {https://arxiv.org/abs/2503.01422} {Sampling-efficient test-time scaling: Self-estimating the best-of-n sampling in early decoding}.
\newblock \emph{Preprint}, arXiv:2503.01422.

\bibitem[{Wei et~al.(2022)Wei, Wang, Schuurmans, Bosma, brian ichter, Xia, Chi, Le, and Zhou}]{wei2022chain}
Jason Wei, Xuezhi Wang, Dale Schuurmans, Maarten Bosma, brian ichter, Fei Xia, Ed~H. Chi, Quoc~V Le, and Denny Zhou. 2022.
\newblock \href {https://openreview.net/forum?id=_VjQlMeSB_J} {Chain of thought prompting elicits reasoning in large language models}.
\newblock In \emph{Advances in Neural Information Processing Systems}.

\bibitem[{Xie et~al.(2024)Xie, Guo, Yu, and Li}]{internal-consistency}
Zhihui Xie, Jizhou Guo, Tong Yu, and Shuai Li. 2024.
\newblock \href {https://openreview.net/forum?id=udZKVMPf3S} {Calibrating reasoning in language models with internal consistency}.
\newblock In \emph{The Thirty-eighth Annual Conference on Neural Information Processing Systems}.

\bibitem[{Xu et~al.(2025{\natexlab{a}})Xu, Guo, Zeng, and Miao}]{xu-etal-2025-softcot}
Yige Xu, Xu~Guo, Zhiwei Zeng, and Chunyan Miao. 2025{\natexlab{a}}.
\newblock \href {https://doi.org/10.18653/v1/2025.acl-long.1137} {{S}oft{C}o{T}: Soft chain-of-thought for efficient reasoning with {LLM}s}.
\newblock In \emph{Proceedings of the 63rd Annual Meeting of the Association for Computational Linguistics (Volume 1: Long Papers)}, pages 23336--23351, Vienna, Austria. Association for Computational Linguistics.

\bibitem[{Xu et~al.(2025{\natexlab{b}})Xu, Guo, Zeng, and Miao}]{xu2025softcottesttimescalingsoft}
Yige Xu, Xu~Guo, Zhiwei Zeng, and Chunyan Miao. 2025{\natexlab{b}}.
\newblock \href {https://arxiv.org/abs/2505.11484} {Softcot++: Test-time scaling with soft chain-of-thought reasoning}.
\newblock \emph{Preprint}, arXiv:2505.11484.

\bibitem[{Yang et~al.(2025)Yang, Li, Yang, Zhang, Hui, Zheng, Yu, Gao, Huang, Lv, Zheng, Liu, Zhou, Huang, Hu, Ge, Wei, Lin, Tang, Yang, Tu, Zhang, Yang, Yang, Zhou, Zhou, Lin, Dang, Bao, Yang, Yu, Deng, Li, Xue, Li, Zhang, Wang, Zhu, Men, Gao, Liu, Luo, Li, Tang, Yin, Ren, Wang, Zhang, Ren, Fan, Su, Zhang, Zhang, Wan, Liu, Wang, Cui, Zhang, Zhou, and Qiu}]{yang2025qwen3technicalreport}
An~Yang, Anfeng Li, Baosong Yang, Beichen Zhang, Binyuan Hui, Bo~Zheng, Bowen Yu, Chang Gao, Chengen Huang, Chenxu Lv, Chujie Zheng, Dayiheng Liu, Fan Zhou, Fei Huang, Feng Hu, Hao Ge, Haoran Wei, Huan Lin, Jialong Tang, and 41 others. 2025.
\newblock \href {https://arxiv.org/abs/2505.09388} {Qwen3 technical report}.
\newblock \emph{Preprint}, arXiv:2505.09388.

\bibitem[{Yao et~al.(2023)Yao, Yu, Zhao, Shafran, Griffiths, Cao, and Narasimhan}]{yao2023tree}
Shunyu Yao, Dian Yu, Jeffrey Zhao, Izhak Shafran, Thomas~L. Griffiths, Yuan Cao, and Karthik~R Narasimhan. 2023.
\newblock \href {https://openreview.net/forum?id=5Xc1ecxO1h} {Tree of thoughts: Deliberate problem solving with large language models}.
\newblock In \emph{Thirty-seventh Conference on Neural Information Processing Systems}.

\bibitem[{Zhou et~al.(2025)Zhou, Yuhao, Li, Yao, Guo, Ma, and Li}]{RPC}
Zhi Zhou, Tan Yuhao, Zenan Li, Yuan Yao, Lan-Zhe Guo, Xiaoxing Ma, and Yu-Feng Li. 2025.
\newblock \href {https://arxiv.org/abs/2502.00511} {Bridging internal probability and self-consistency for effective and efficient llm reasoning}.
\newblock \emph{Preprint}, arXiv:2502.00511.

\bibitem[{Zhu et~al.(2025)Zhu, Shen, Zhao, and Zou}]{path-consistency-SC}
Jiace Zhu, Yingtao Shen, Jie Zhao, and An~Zou. 2025.
\newblock \href {https://arxiv.org/abs/2409.01281} {Path-consistency: Prefix enhancement for efficient inference in llm}.
\newblock \emph{Preprint}, arXiv:2409.01281.

\end{thebibliography}

\appendix

\onecolumn

\section{Semantic similarity trajectories for Qwen2.5-32B}
\label{app:sim}
\begin{figure*}[htbp]
  \centering
  \begin{subfigure}[t]{0.98\textwidth}
    \centering
    \includegraphics[width=\textwidth]{legend.png}
  \end{subfigure}
  \hfill
  
  \begin{subfigure}[t]{0.47\textwidth}
    \centering
    \includegraphics[width=\textwidth]{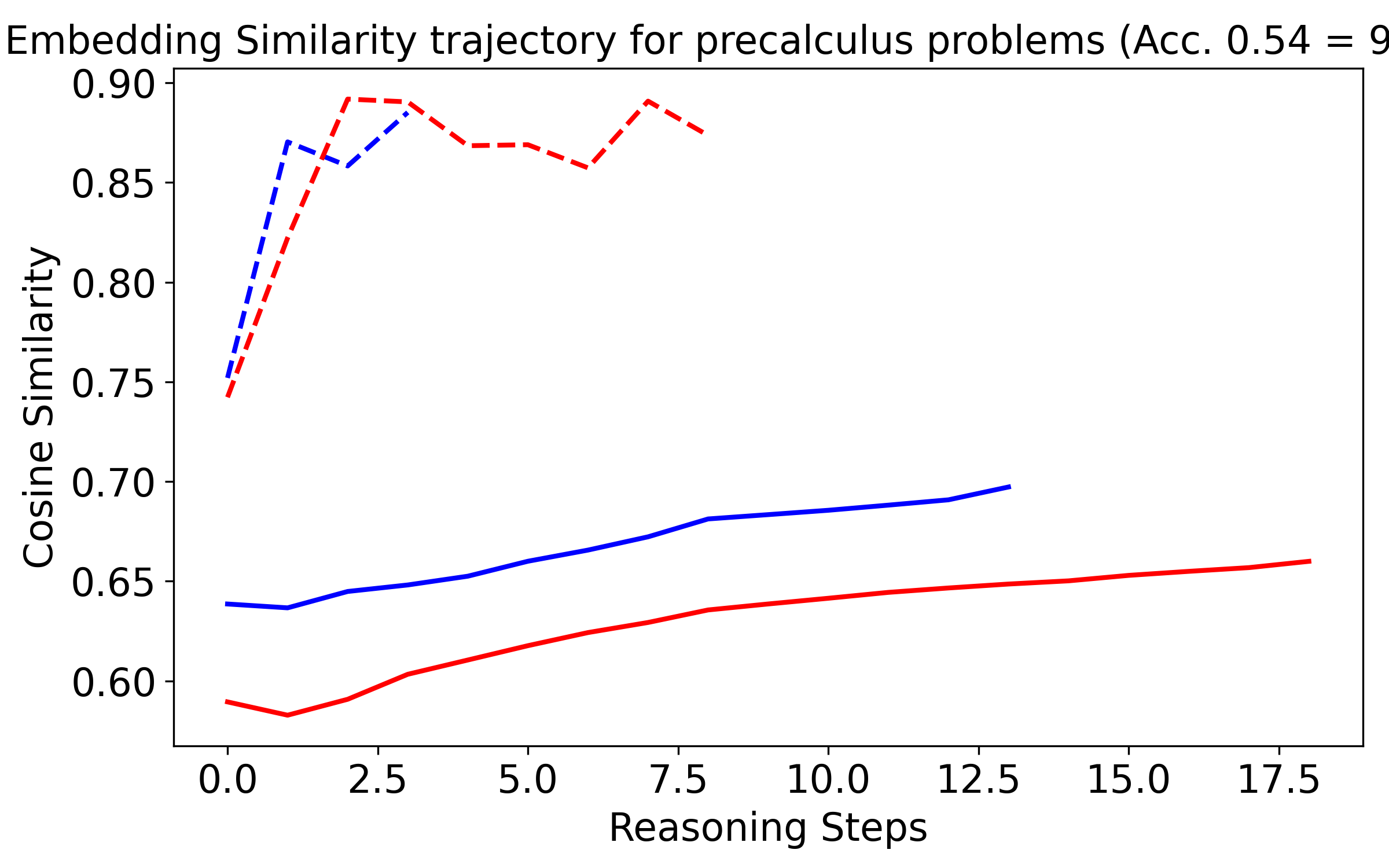}
    \caption{GPT-4o vs. Human Reasoning on Precalculus. }
  \end{subfigure}
  \hfill
  \begin{subfigure}[t]{0.47\textwidth}
    \centering
    \includegraphics[width=\textwidth]{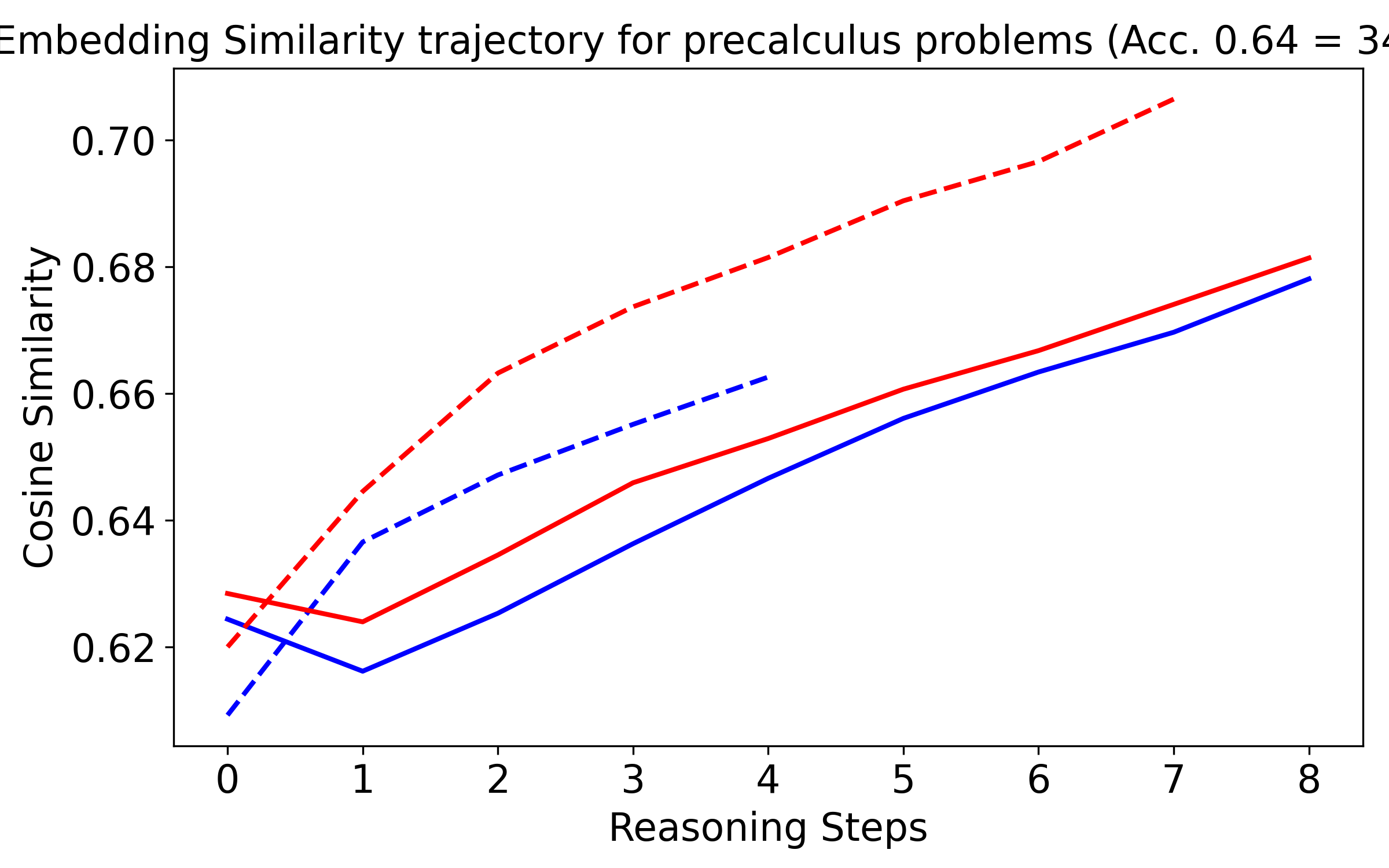}
    \caption{Qwen2.5-32B vs. Human Reasoning on Precalculus. }
  \end{subfigure}
  \hfill
  \begin{subfigure}[t]{0.47\textwidth}
    \centering
    \includegraphics[width=\textwidth]{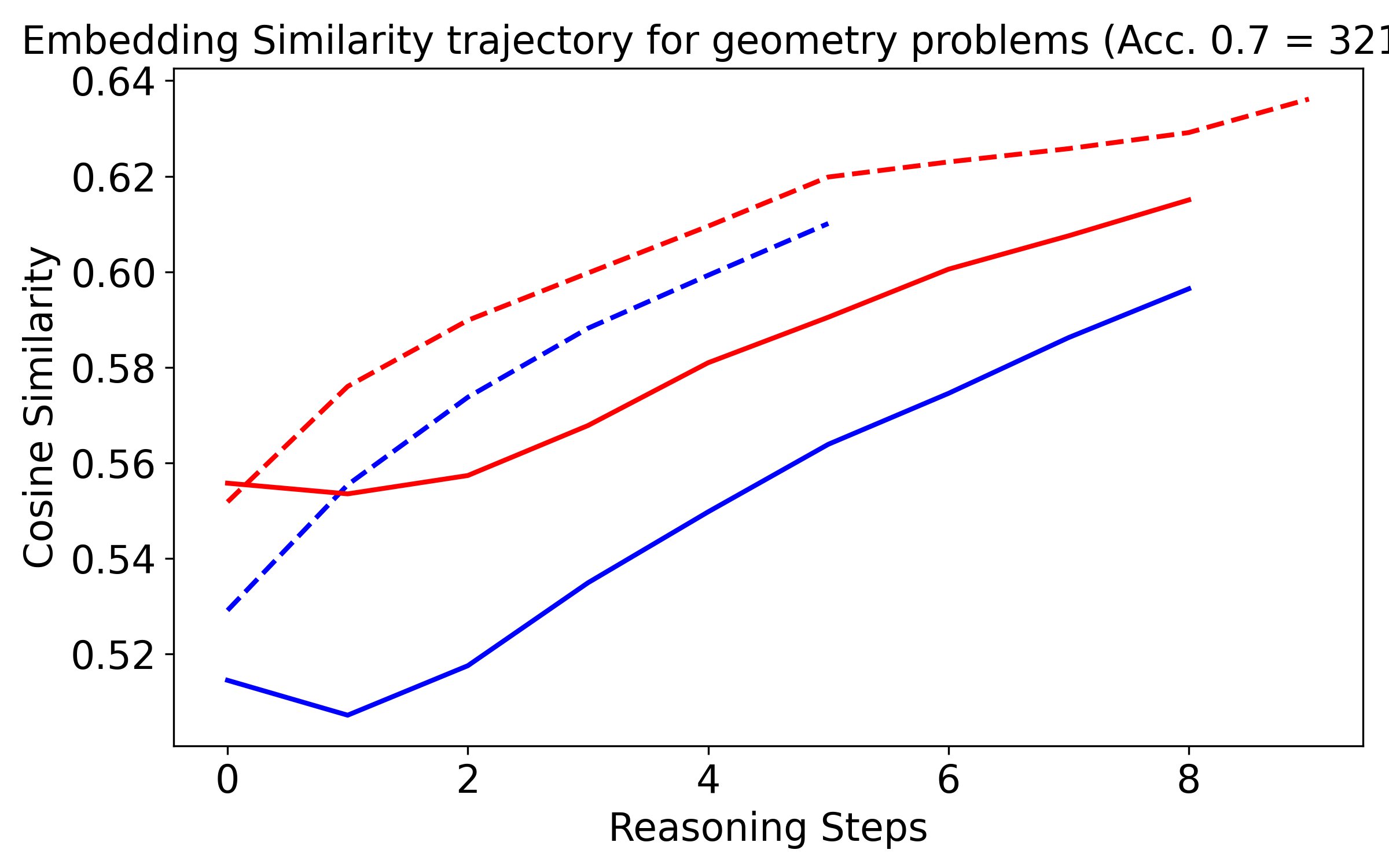}
    \caption{Qwen2.5-32B vs. Human Reasoning on Geometry.}
  \end{subfigure}
  \hfill
  \begin{subfigure}[t]{0.47\textwidth}
    \centering
    \includegraphics[width=\textwidth]{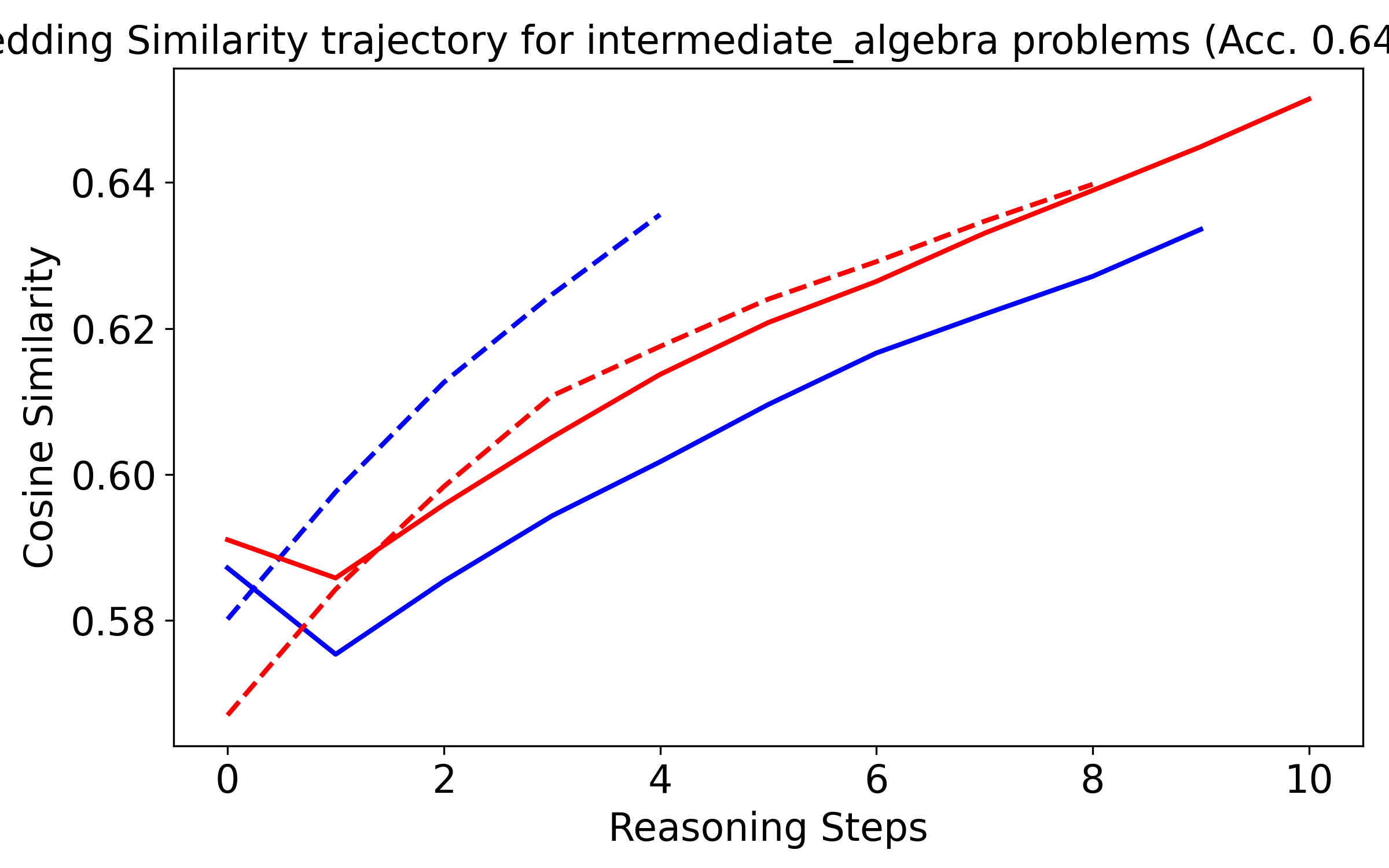}
    \caption{Qwen2.5-32B vs. Human on Intermediate Algebra.}
  \end{subfigure}
  \hfill
  \begin{subfigure}[t]{0.47\textwidth}
    \centering
    \includegraphics[width=\textwidth]{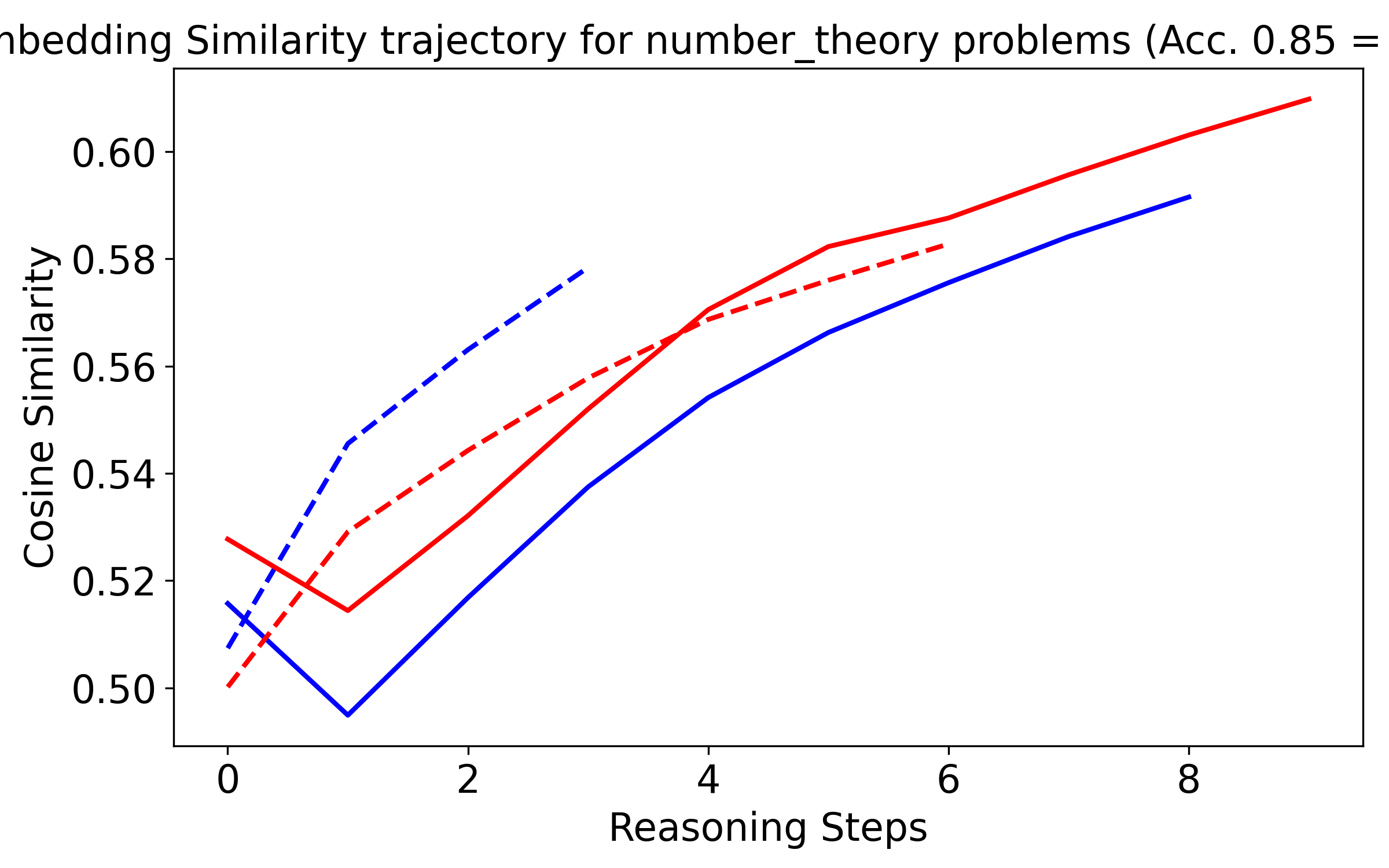}
    \caption{Qwen2.5-32B vs. Human Reasoning on Number Theory.}
  \end{subfigure}
  \hfill
  \begin{subfigure}[t]{0.47\textwidth}
    \centering
    \includegraphics[width=\textwidth]{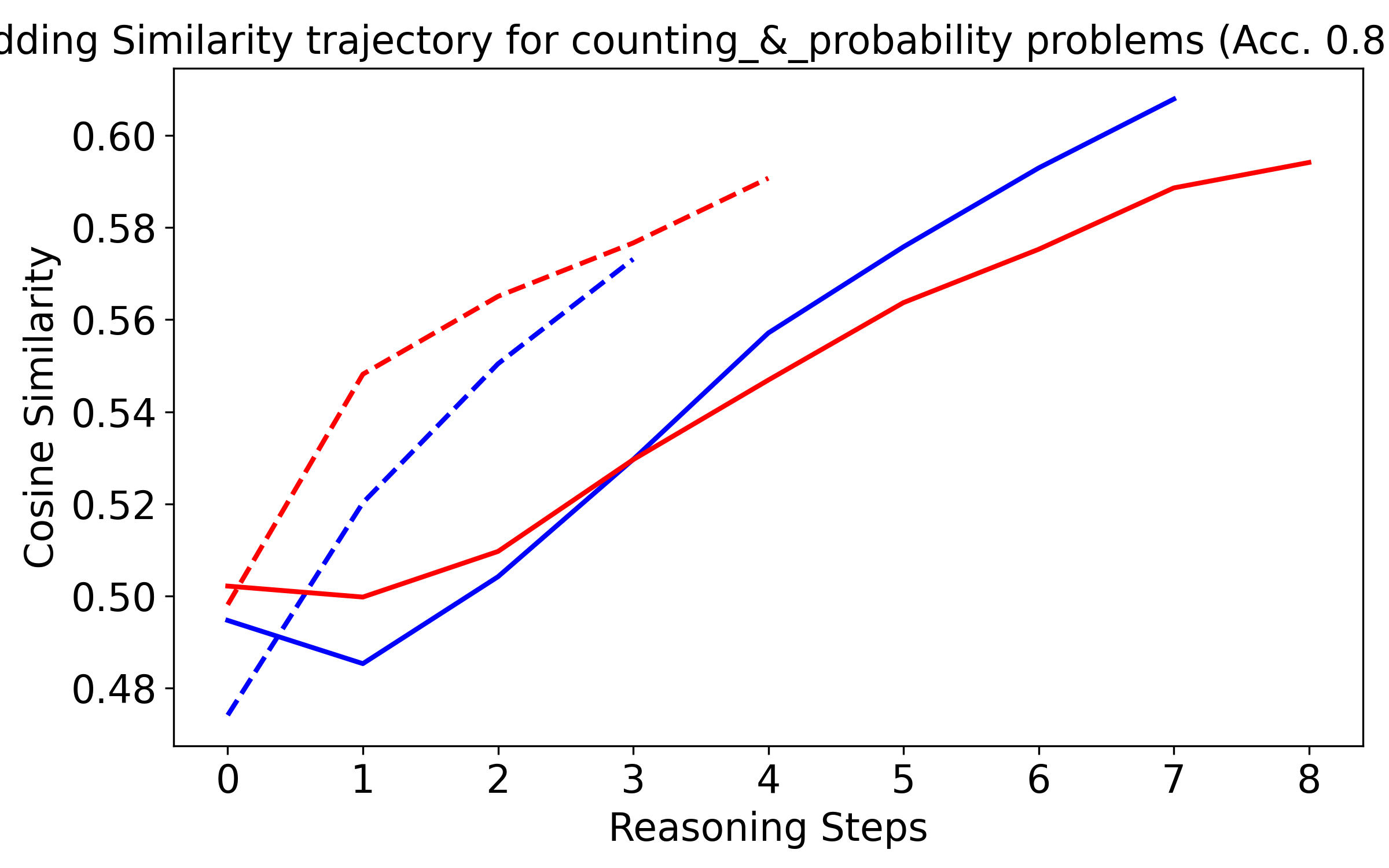}
    \caption{Qwen2.5-32B vs. Human on Counting \& Probability.}
  \end{subfigure}
  \hfill
  \begin{subfigure}[t]{0.47\textwidth}
    \centering
    \includegraphics[width=\textwidth]{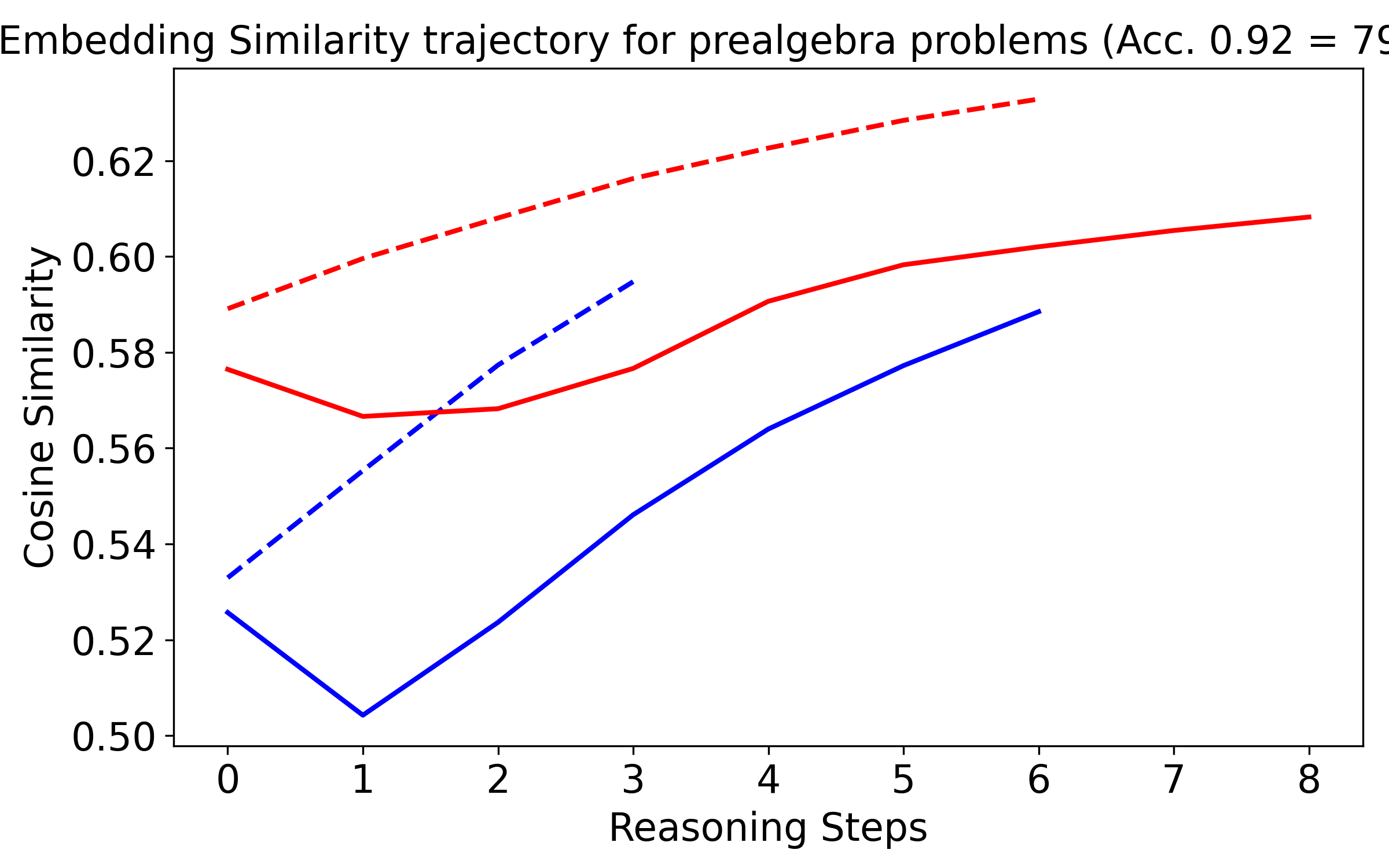}
    \caption{Qwen2.5-32B vs. Human Reasoning on Prealgebra.}
  \end{subfigure}
  \hfill
  \begin{subfigure}[t]{0.47\textwidth}
    \centering
    \includegraphics[width=\textwidth]{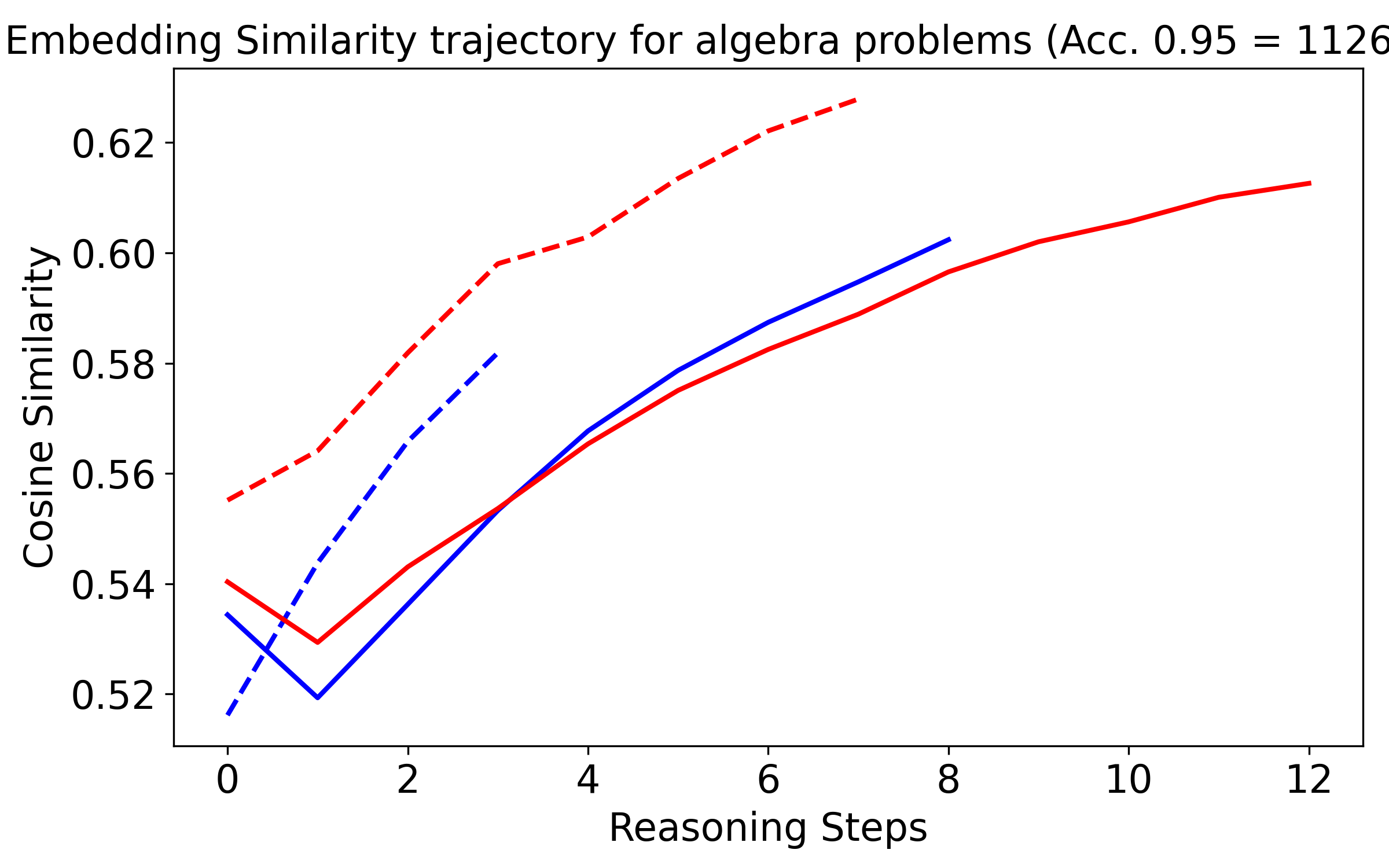}
    \caption{Qwen2.5-32B vs. Human Reasoning on Algebra.}
  \end{subfigure}
  \caption{Cosine-similarity trajectories between each reasoning step ($C_{\leq k}$) and the ground-truth answer span ($Y$) for Qwen2.5-32B. Correct and incorrect chains show no consistent separation, indicating little to no correlation between similarity and accuracy.}
  \label{fig:cosine-sim-qwen}
\end{figure*}

We study whether cosine similarity between the evolving context and the gold answer span separates correct from incorrect reasoning chains (Figure~\ref{fig:cosine-sim-qwen}). At each reasoning step $k$, we form $C_k=[X; Z_{\le k}]$ and compute mean-pooled token representations for $C_k$ and for the answer span $Y=(y_1,\dots,y_M)$. The cosine similarity at step-$k$ is
$s_k=\cos\!\big(\bar{\mathbf e}(C_k),\,\bar{\mathbf e}(Y)\big)$.

We plot $\{s_k\}_{k=1}^K$ for correct and incorrect chains using the same cubic-spline interpolation to reveal general trends. We find that (1) Similarity trajectories show limited separability between correct and incorrect chains, indicating weak predictive power for accuracy. (2) Nonetheless, $s_k$ generally increases with $k$: as reasoning unfolds, the context representation moves closer to the answer-span representation, though the answer is a short numeric expression. This suggests that the semantics of the natural-language derivation for a math problem become progressively closer to the target number in the model’s representation space, although this alignment alone does not determine correctness.

\section{Cross-entropy trajectories for Qwen2.5-32B}
\label{app:cross}
\begin{figure*}[htbp]
  \centering
  \begin{subfigure}[t]{0.98\textwidth}
    \centering
    \includegraphics[width=\textwidth]{legend.png}
  \end{subfigure}
  \hfill
  
  \begin{subfigure}[t]{0.47\textwidth}
    \centering
    \includegraphics[width=\textwidth]{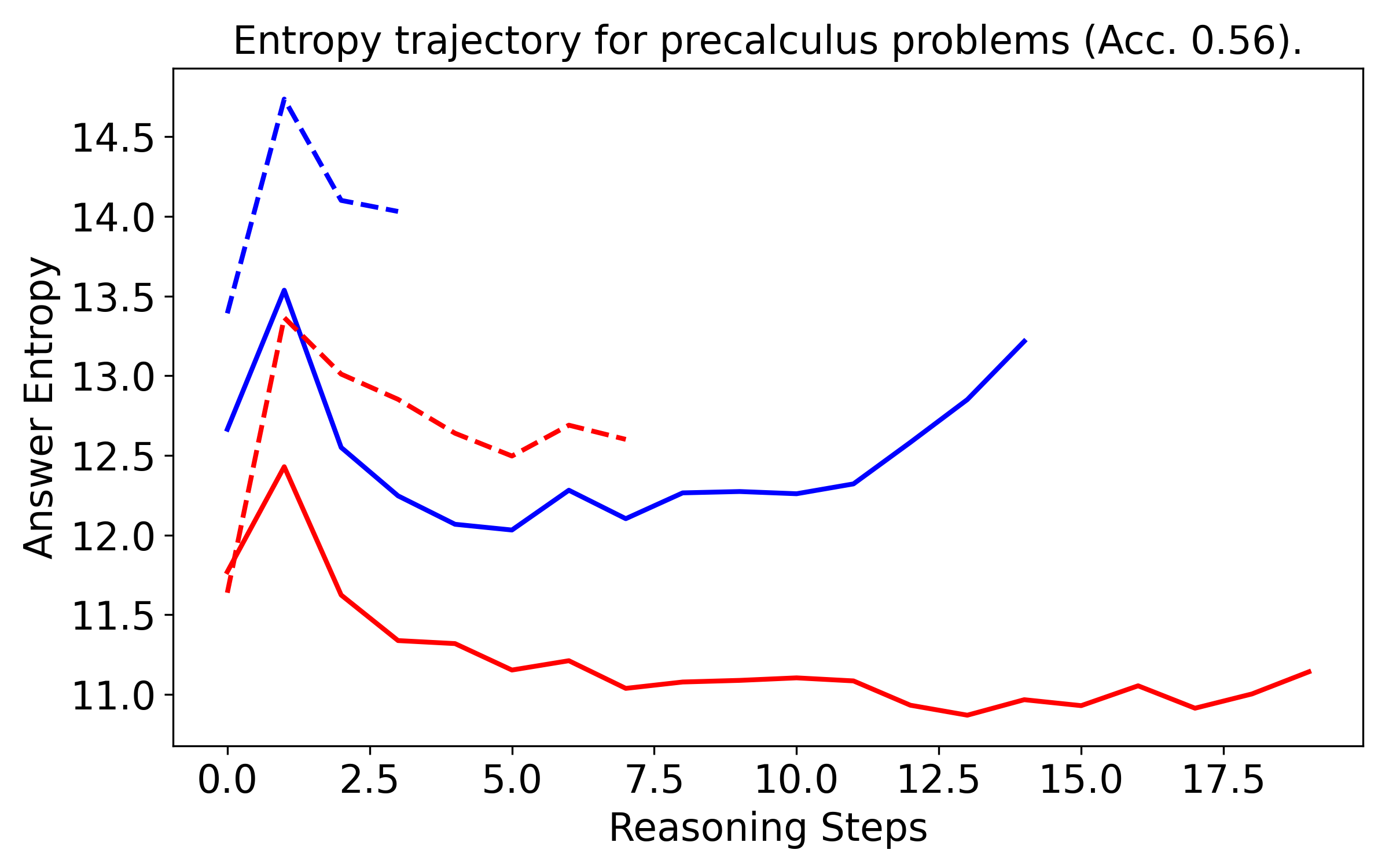}
    \caption{GPT-4o vs. Human Reasoning on Precalculus. }
  \end{subfigure}
  \hfill
  \begin{subfigure}[t]{0.47\textwidth}
    \centering
    \includegraphics[width=\textwidth]{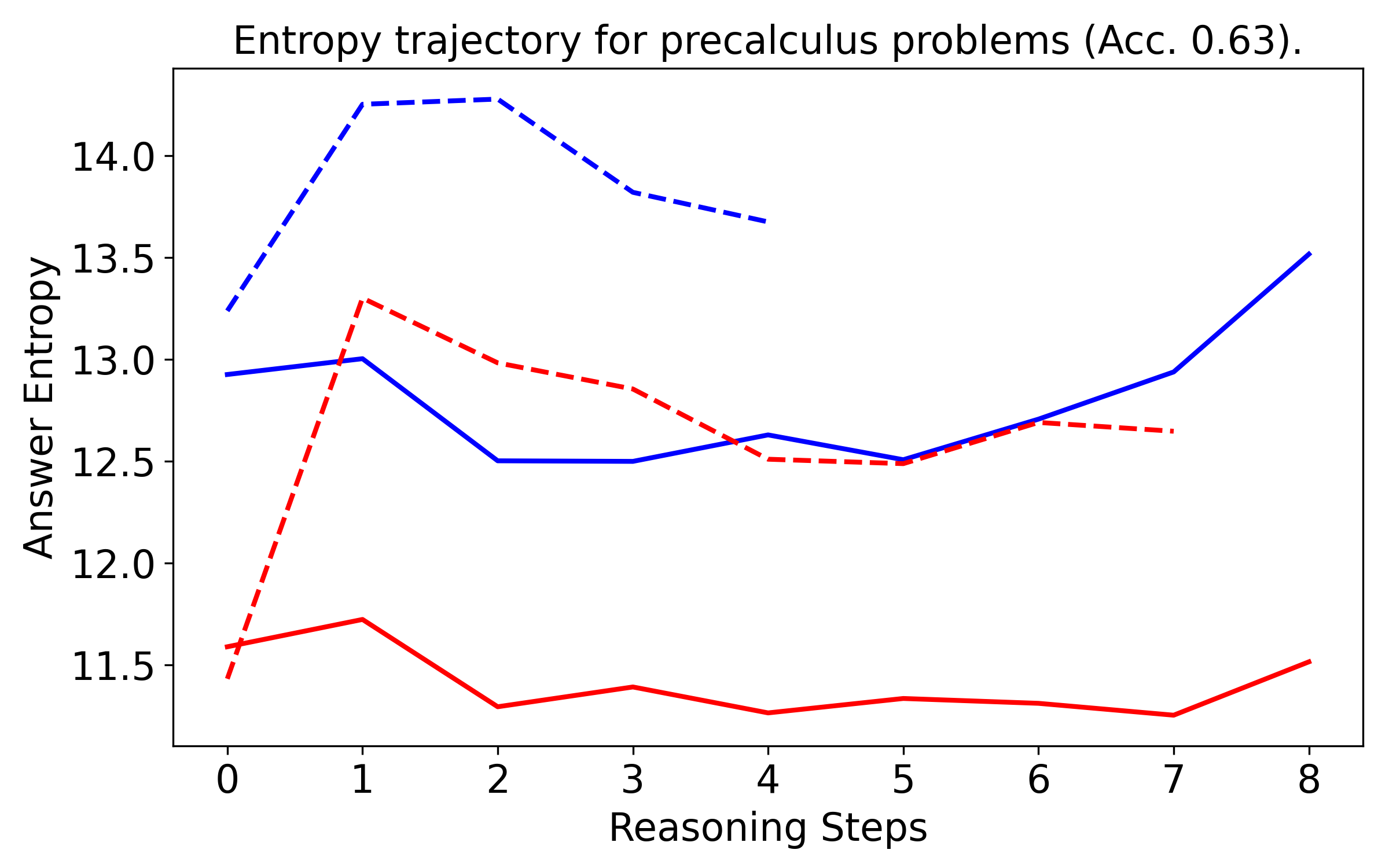}
    \caption{Qwen2.5-32B vs. Human Reasoning on Precalculus. }
  \end{subfigure}
  \hfill
  \begin{subfigure}[t]{0.47\textwidth}
    \centering
    \includegraphics[width=\textwidth]{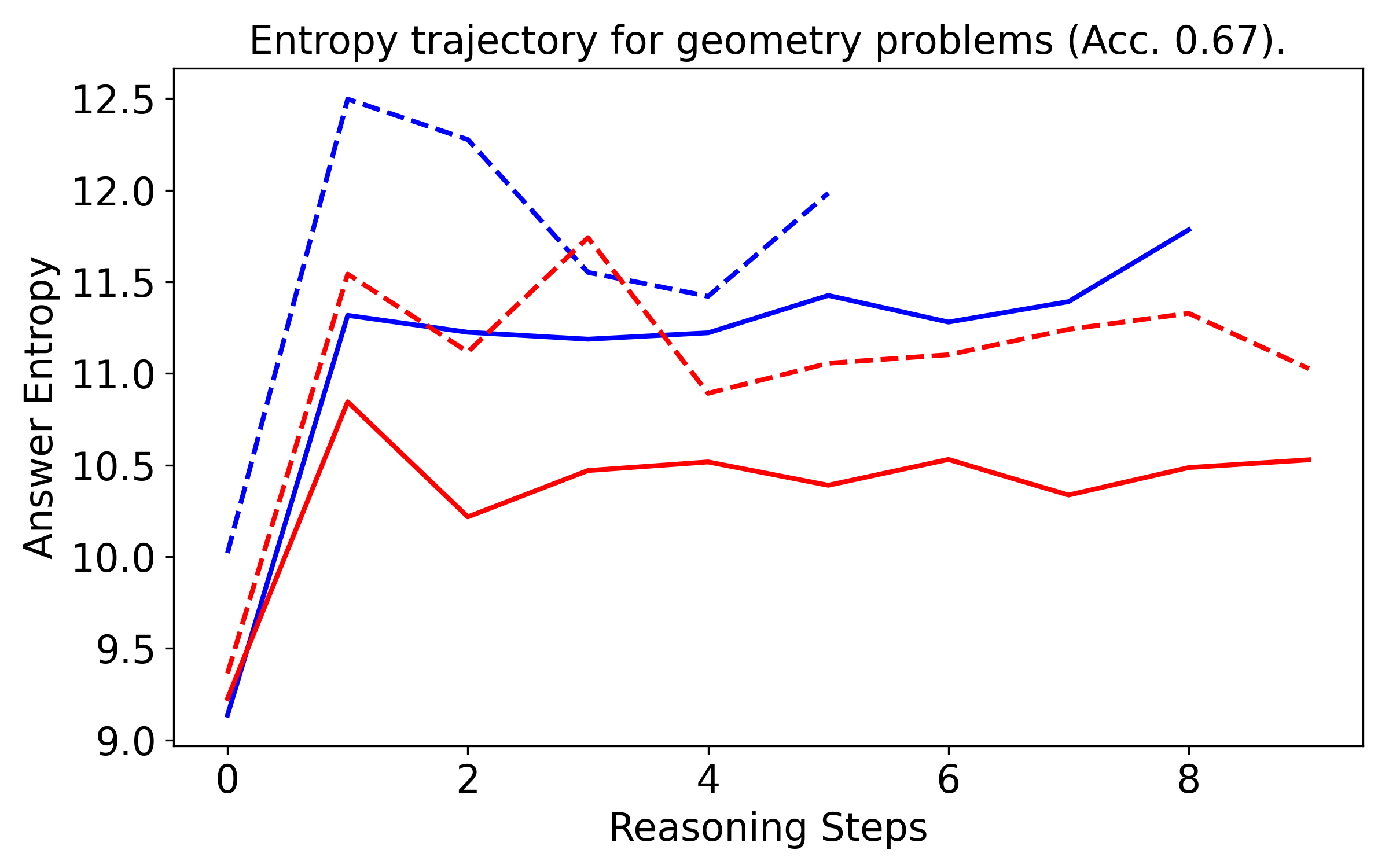}
    \caption{Qwen2.5-32B vs. Human Reasoning on Geometry.}
  \end{subfigure}
  \hfill
  \begin{subfigure}[t]{0.47\textwidth}
    \centering
    \includegraphics[width=\textwidth]{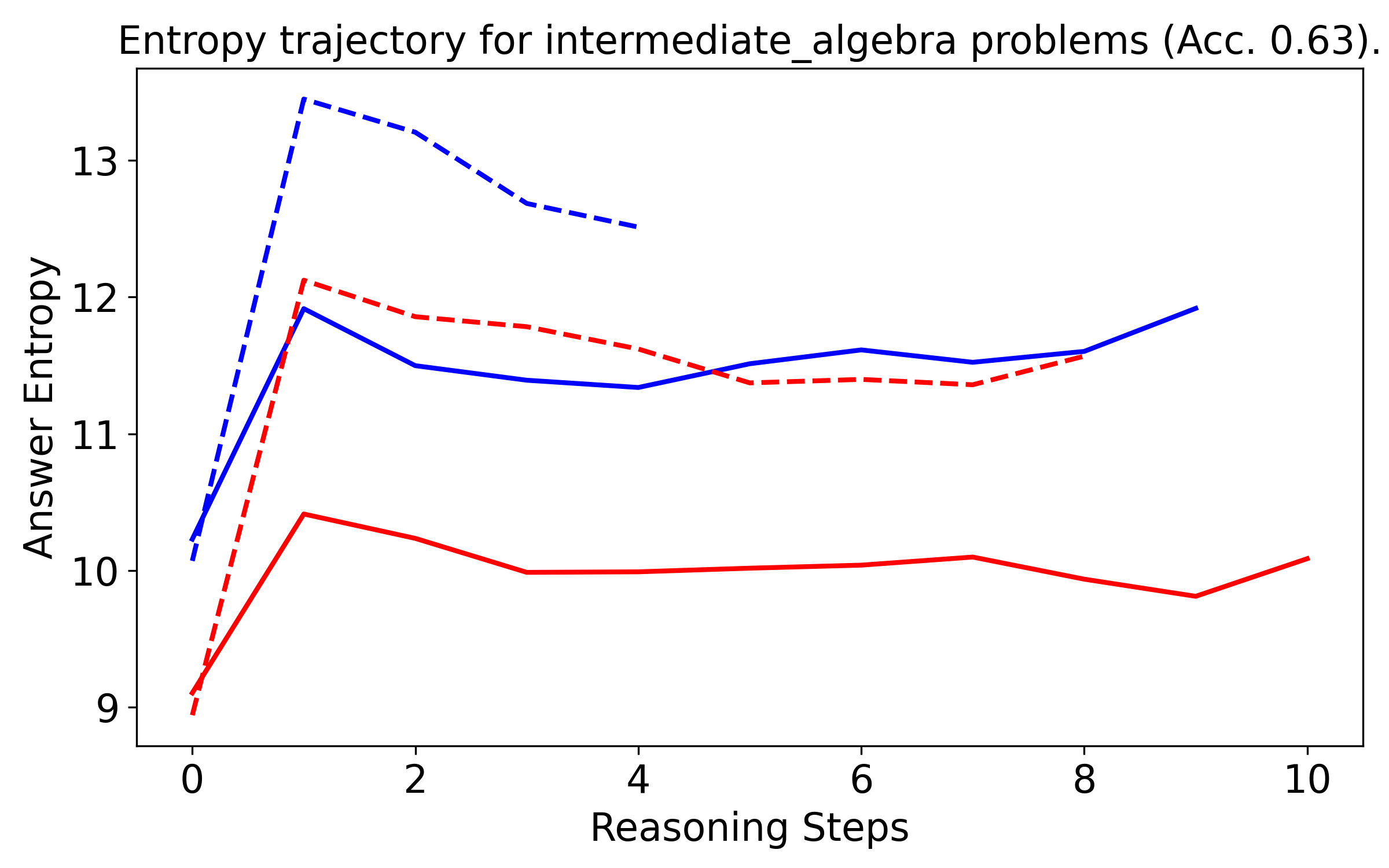}
    \caption{Qwen2.5-32B vs. Human on Intermediate Algebra.}
  \end{subfigure}
  \hfill
  \begin{subfigure}[t]{0.47\textwidth}
    \centering
    \includegraphics[width=\textwidth]{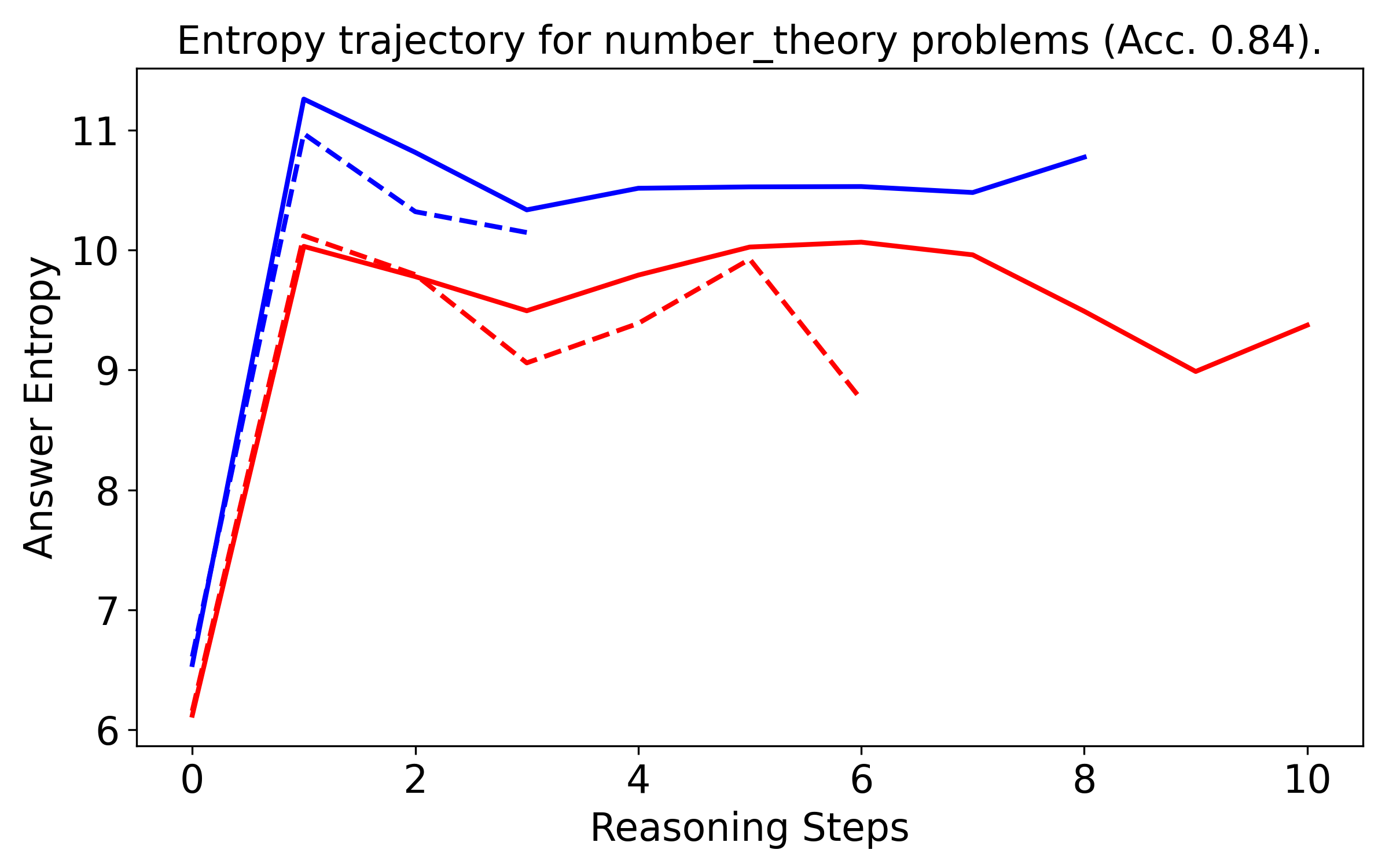}
    \caption{Qwen2.5-32B vs. Human Reasoning on Number Theory.}
  \end{subfigure}
  \hfill
  \begin{subfigure}[t]{0.47\textwidth}
    \centering
    \includegraphics[width=\textwidth]{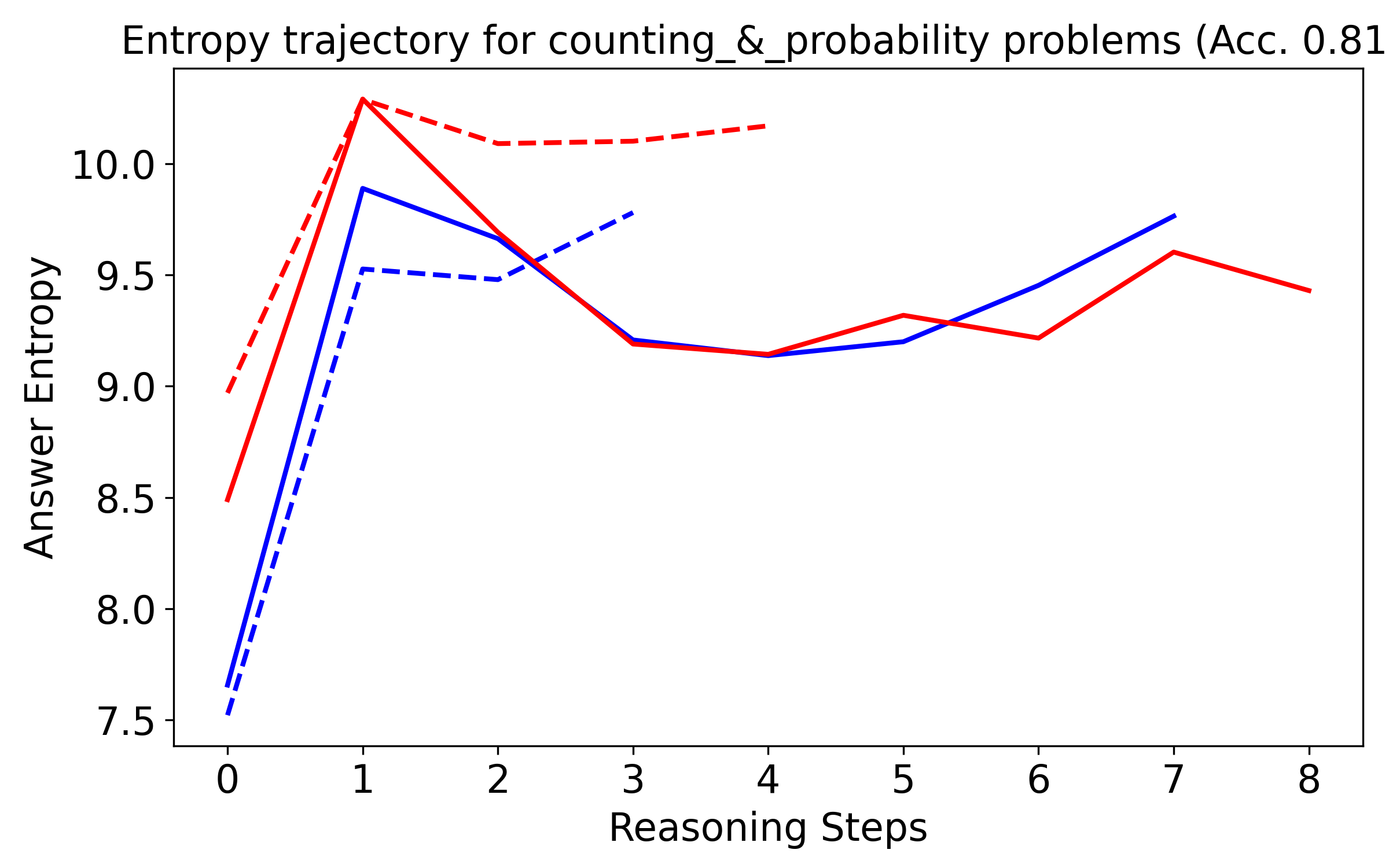}
    \caption{Qwen2.5-32B vs. Human on Counting \& Probability.}
  \end{subfigure}
  \hfill
  \begin{subfigure}[t]{0.47\textwidth}
    \centering
    \includegraphics[width=\textwidth]{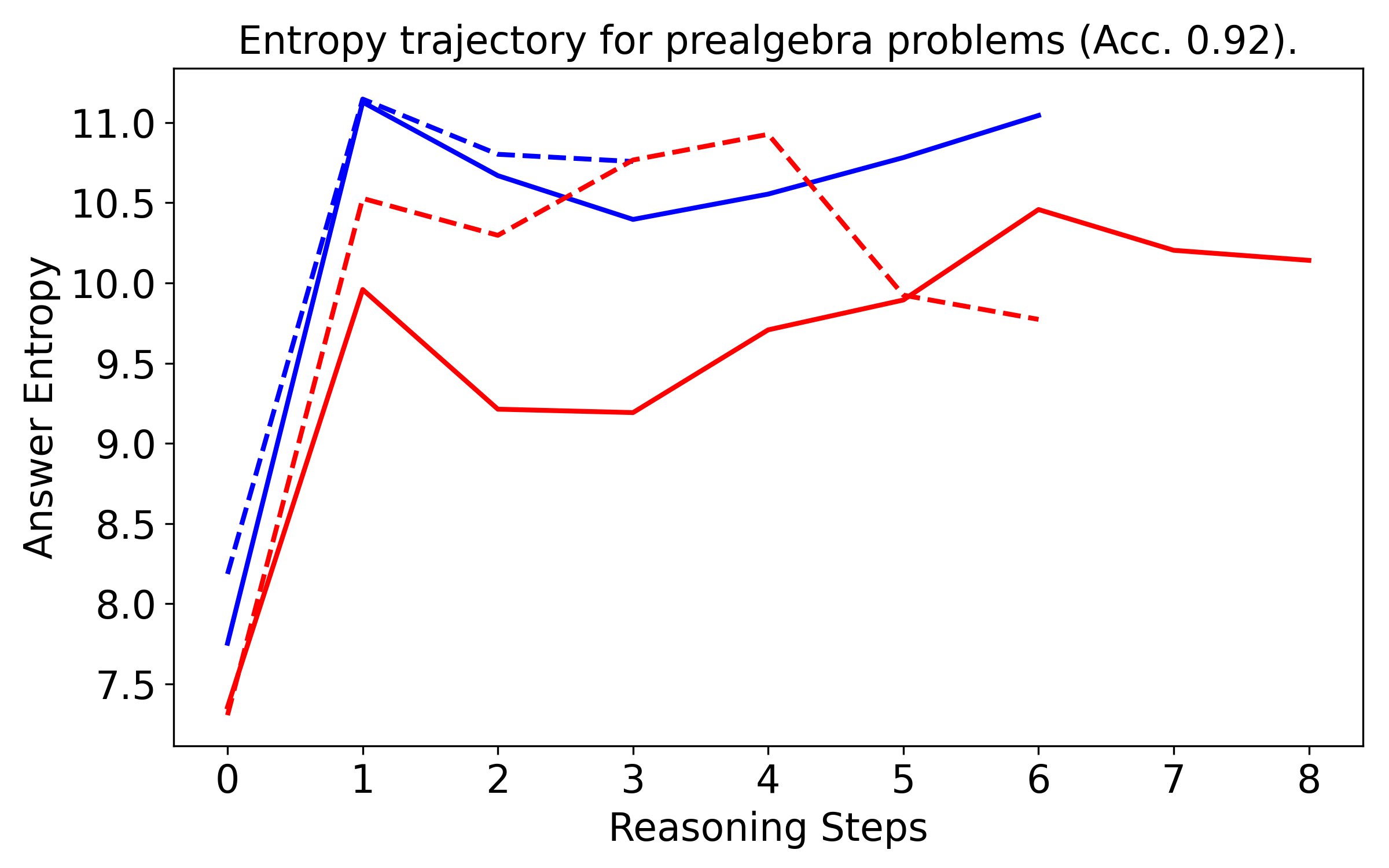}
    \caption{Qwen2.5-32B vs. Human Reasoning on Prealgebra.}
  \end{subfigure}
  \hfill
  \begin{subfigure}[t]{0.47\textwidth}
    \centering
    \includegraphics[width=\textwidth]{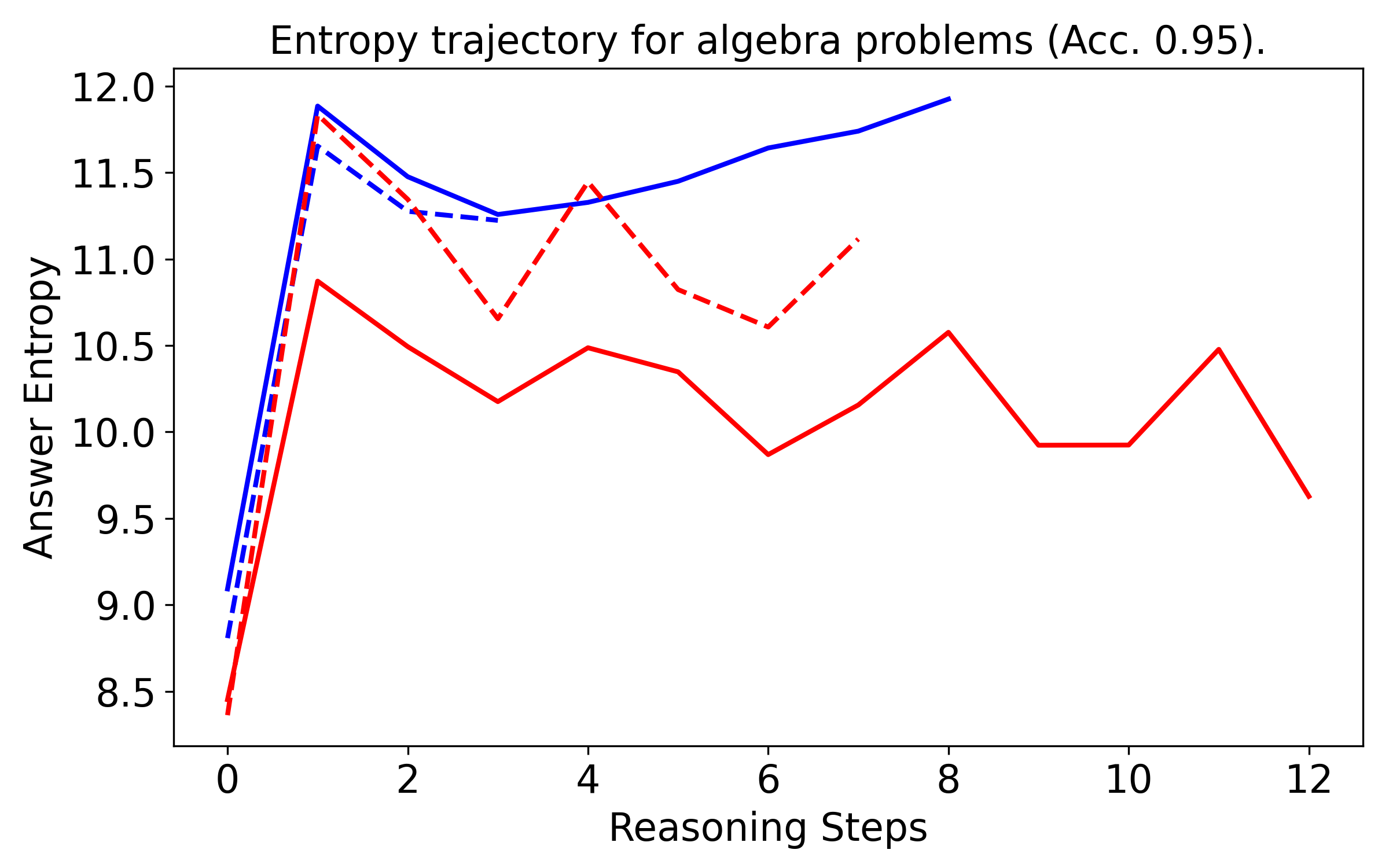}
    \caption{Qwen2.5-32B vs. Human Reasoning on Algebra.}
  \end{subfigure}
  \caption{Cross-Entropy variation across reasoning steps for Qwen2.5-32B. Trajectories generally plateau and show no stable correlation with accuracy. A notable finding is that LLMs tend to be more confident on incorrect chains than on correct chains, suggesting that confidence alone may not lead to an accurate answer. The y-axis values are the \textit{average} cross-entropy over the answer span (same tokens in the span can lead to higher negative log-likelihood). }
  \label{fig:cross-entropy-qwen}
\end{figure*}

We also report the cross-entropy trajectory during reasoning (Figure~\ref{fig:cross-entropy-qwen}).  
At each reasoning step $k$, we form $C_k=[X; Z_{\leq k}]$ and compute:
\[
\text{CE}(Y \mid C_k) = \frac{1}{|Y|} \sum_{t=1}^{|Y|} - \log p(y_t \mid C_k, y_{<t}).
\]
The key difference between cross-entropy and entropy is that cross-entropy only aggregates $-\log p(y_t)$ for the ground-truth tokens, while entropy averages self-information $-p(v)\log p(v)$ across all $v \in \mathcal{V}$.  
Other commonly used metrics are directly derived from cross-entropy: \textit{perplexity} is its exponential, and \textit{confidence} corresponds to $p(y_t \mid C_k, y_{<t})$ for the ground-truth token.

\noindent Figure~\ref{fig:cross-entropy-qwen} shows that cross-entropy tends to plateau as reasoning unfolds, both for humans and LLMs.  
Interestingly, we do not observe a consistent difference between correct and incorrect chains.  
Moreover, LLMs often display \textit{lower} cross-entropy (i.e., higher confidence) on incorrect chains.  
This suggests a weak link between confidence and correctness: higher confidence does not reliably imply correctness, raising concerns about confidence-based reasoning evaluation.

\section{Examples for human solution and LLM solution}
\label{app:example}
Humans figure out the answer space early; LLMs take it step by step.
Here, we illustrate a \textcolor{blue}{Number Theory} problem reasoning process below, showing the strategic difference in humans and LLMs. This difference is reflected in the entropy trajectories in Figure \ref{fig:overall} and we discussed this insight in Section \ref{sec:results}.
\begin{tcolorbox}[title=Number Theory Problem]
A positive multiple of 45 less than 1000 is randomly selected. What is the probability that it is a two-digit integer? Express your answer as a common fraction.
\end{tcolorbox}

Below is the \textcolor{blue}{Human Solution}:
\vspace{0.2cm}
\hrule
\vspace{0.2cm}
The positive multiples of 45 are
\[
45, 90, 135, \ldots, 990 = 1\cdot45, 2\cdot45, 3\cdot45, \ldots, 22\cdot45.
\]
There are 22 multiples on this list. Every positive multiple of 45 less than 1000 is either a two-digit integer or a three-digit integer. Out of the \(99 - 10 + 1 = 90\) two-digit integers, \(45\) and \(90\) are multiples of 45. Therefore, the probability that the selected multiple of 45 has two digits is \(2/22 = \boxed{\frac{1}{11}}\).
\vspace{0.2cm}
\hrule
\vspace{0.2cm}
Below is the \textcolor{blue}{Solution generated by Qwen2.5-34B}:
\vspace{0.2cm}
\hrule
\vspace{0.2cm}
Let's solve this math problem step by step. Write the answer in \boxed{}.

Problem: A positive multiple of 45 less than 1000 is randomly selected. What is the probability that it is a two-digit integer? Express your answer as a common fraction.

Solution: To determine the probability that a randomly selected positive multiple of 45 less than 1000 is a two-digit integer, we need to follow these steps:

1. \textbf{Identify the range of multiples of 45 less than 1000:}
   \begin{itemize}
     \item The smallest multiple of 45 is \(45 \times 1 = 45\).
     \item The largest multiple of 45 less than 1000 can be found by dividing 999 by 45 and taking the integer part:
     \[
     \left\lfloor \frac{999}{45} \right\rfloor = \left\lfloor 22.2 \right\rfloor = 22
     \]
     \item Therefore, the largest multiple of 45 less than 1000 is \(45 \times 22 = 990\).
   \end{itemize}

2. \textbf{Count the total number of multiples of 45 less than 1000:}
   \begin{itemize}
     \item These multiples form an arithmetic sequence: \(45, 90, 135, \ldots, 990\).
     \item The number of terms in this sequence is given by:
     \[
     n = 22 - 1 + 1 = 22
     \]
   \end{itemize}

3. \textbf{Identify the two-digit multiples of 45:}
   \begin{itemize}
     \item The smallest two-digit multiple of 45 is \(45 \times 1 = 45\).
     \item The largest two-digit multiple of 45 is \(45 \times 2 = 90\).
     \item The two-digit multiples of 45 are \(45\) and \(90\).
   \end{itemize}

4. \textbf{Count the number of two-digit multiples of 45:}
   \begin{itemize}
     \item There are 2 two-digit multiples of 45.
   \end{itemize}

5. \textbf{Calculate the probability:}
   \begin{itemize}
     \item The probability is the ratio of the number of two-digit multiples to the total number of multiples of 45 less than 1000:
     \[
     \text{Probability} = \frac{\text{Number of two-digit multiples}}{\text{Total number of multiples}} = \frac{2}{22} = \frac{1}{11}
     \]
   \end{itemize}

Thus, the probability that a randomly selected positive multiple of 45 less than 1000 is a two-digit integer is \(\boxed{\frac{1}{11}}\).
\vspace{0.2cm}
\hrule
\vspace{0.2cm}

\begin{figure*}[htbp]
  \centering
  \begin{subfigure}[t]{0.98\textwidth}
    \centering
    \includegraphics[width=\textwidth]{legend.png}
  \end{subfigure}
  \hfill
  
  \begin{subfigure}[t]{0.47\textwidth}
    \centering
    \includegraphics[width=\textwidth]{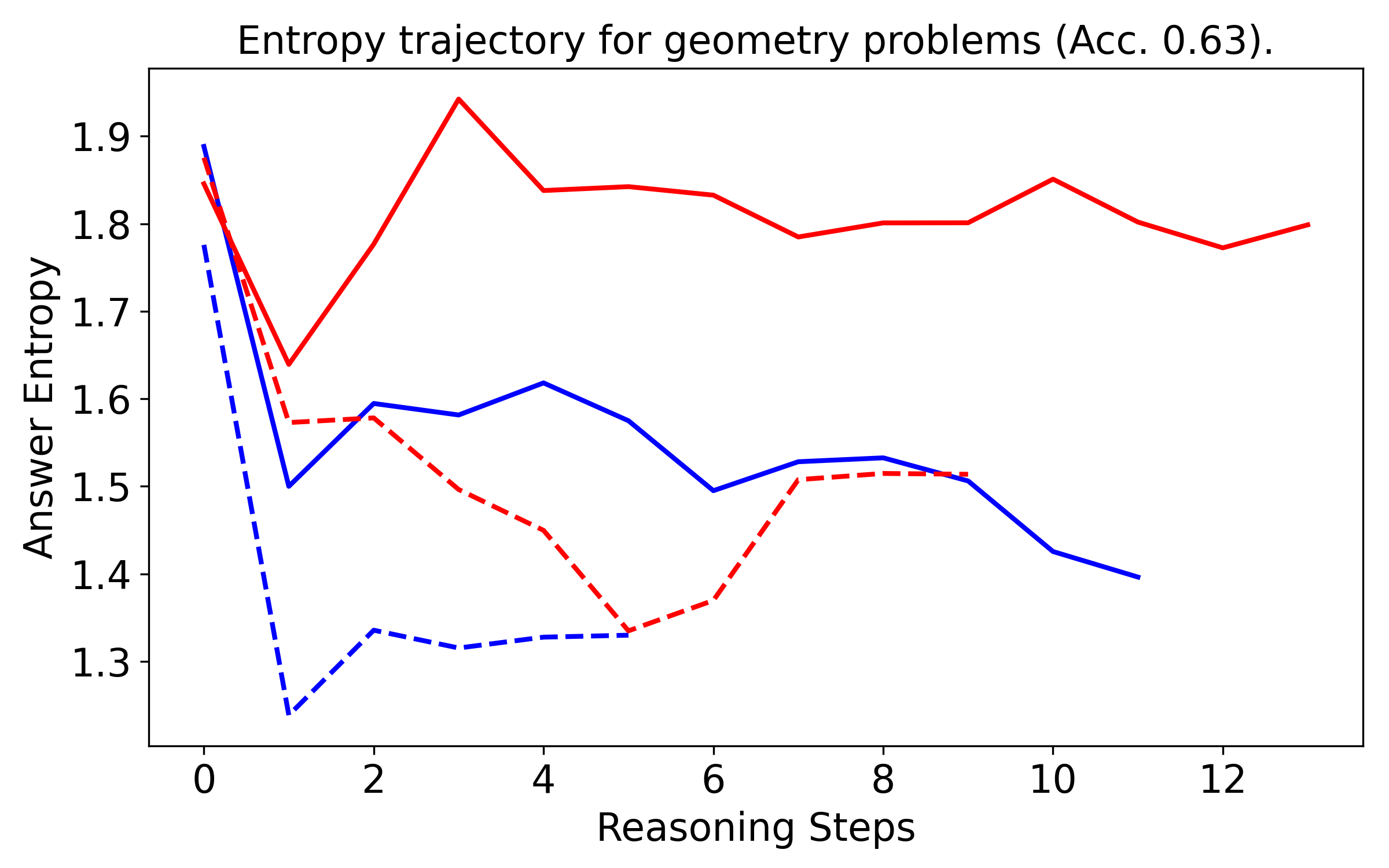}
    \caption{GPT-4o vs. Human Reasoning on Geometry.}
  \end{subfigure}
  \hfill
  \begin{subfigure}[t]{0.47\textwidth}
    \centering
    \includegraphics[width=\textwidth]{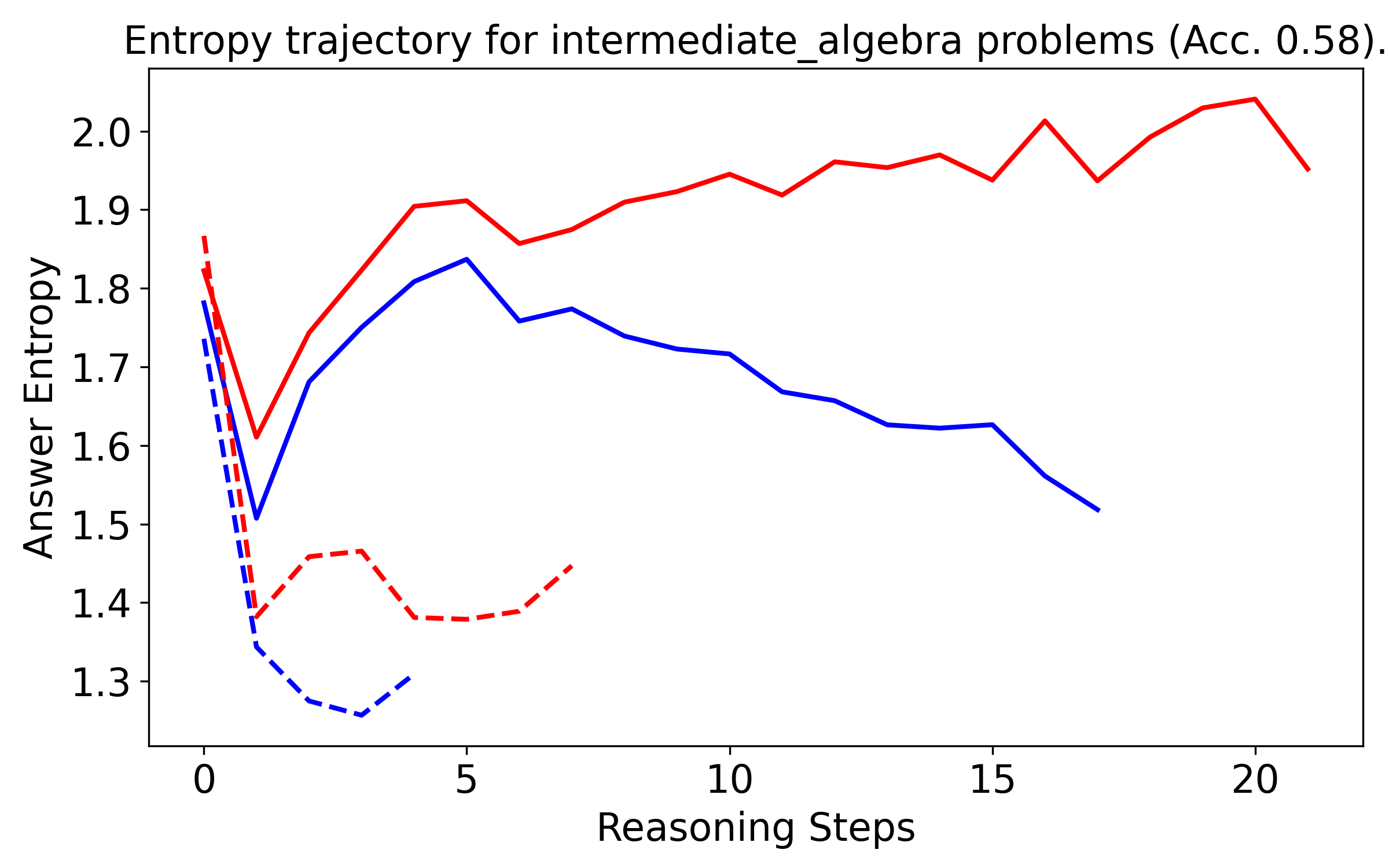}
    \caption{GPT-4o vs. Human on Intermediate Algebra.}
  \end{subfigure}
  \hfill
  \begin{subfigure}[t]{0.47\textwidth}
    \centering
    \includegraphics[width=\textwidth]{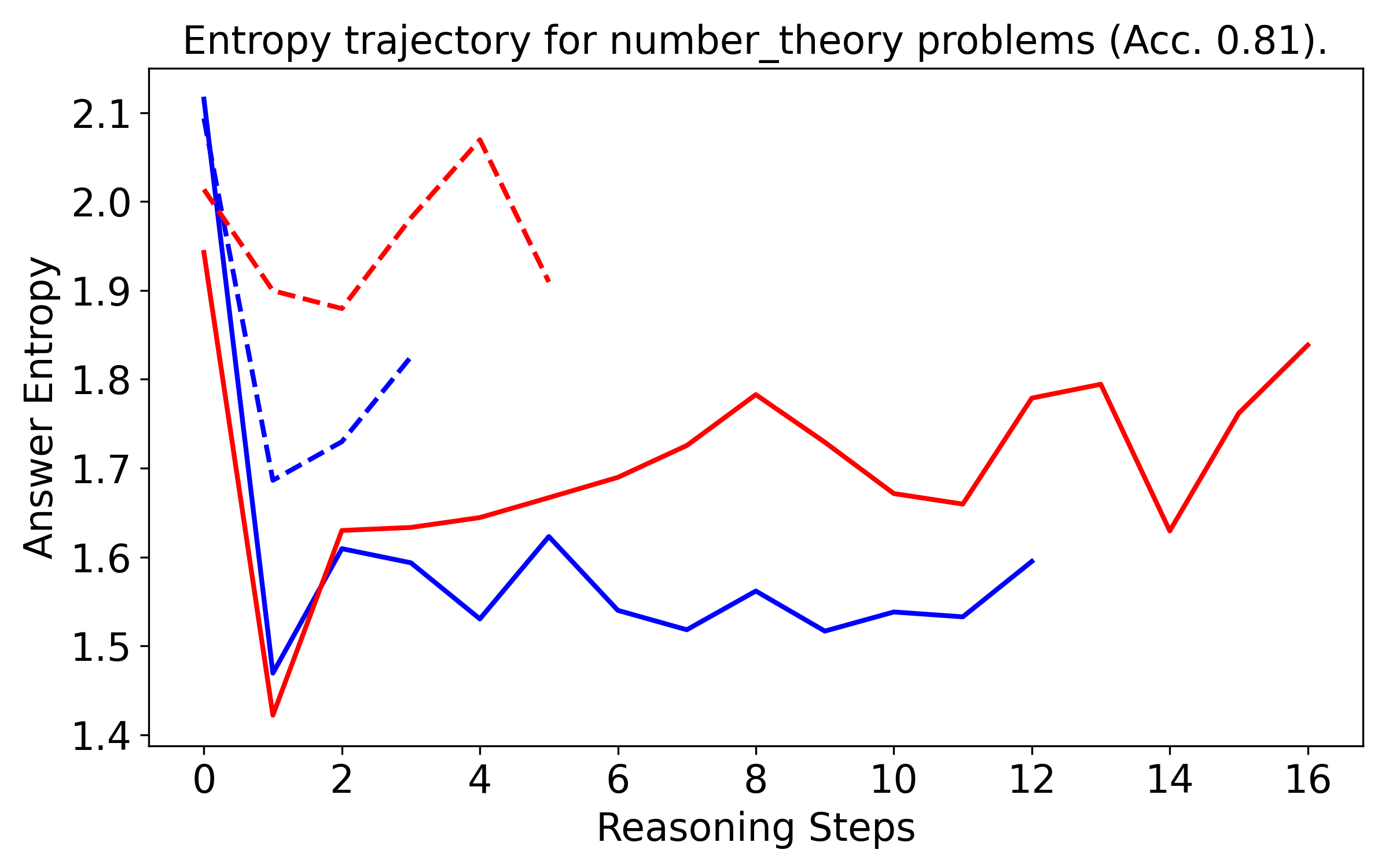}
    \caption{GPT-4o vs. Human Reasoning on Number Theory.}
  \end{subfigure}
  \hfill
  \begin{subfigure}[t]{0.47\textwidth}
    \centering
    \includegraphics[width=\textwidth]{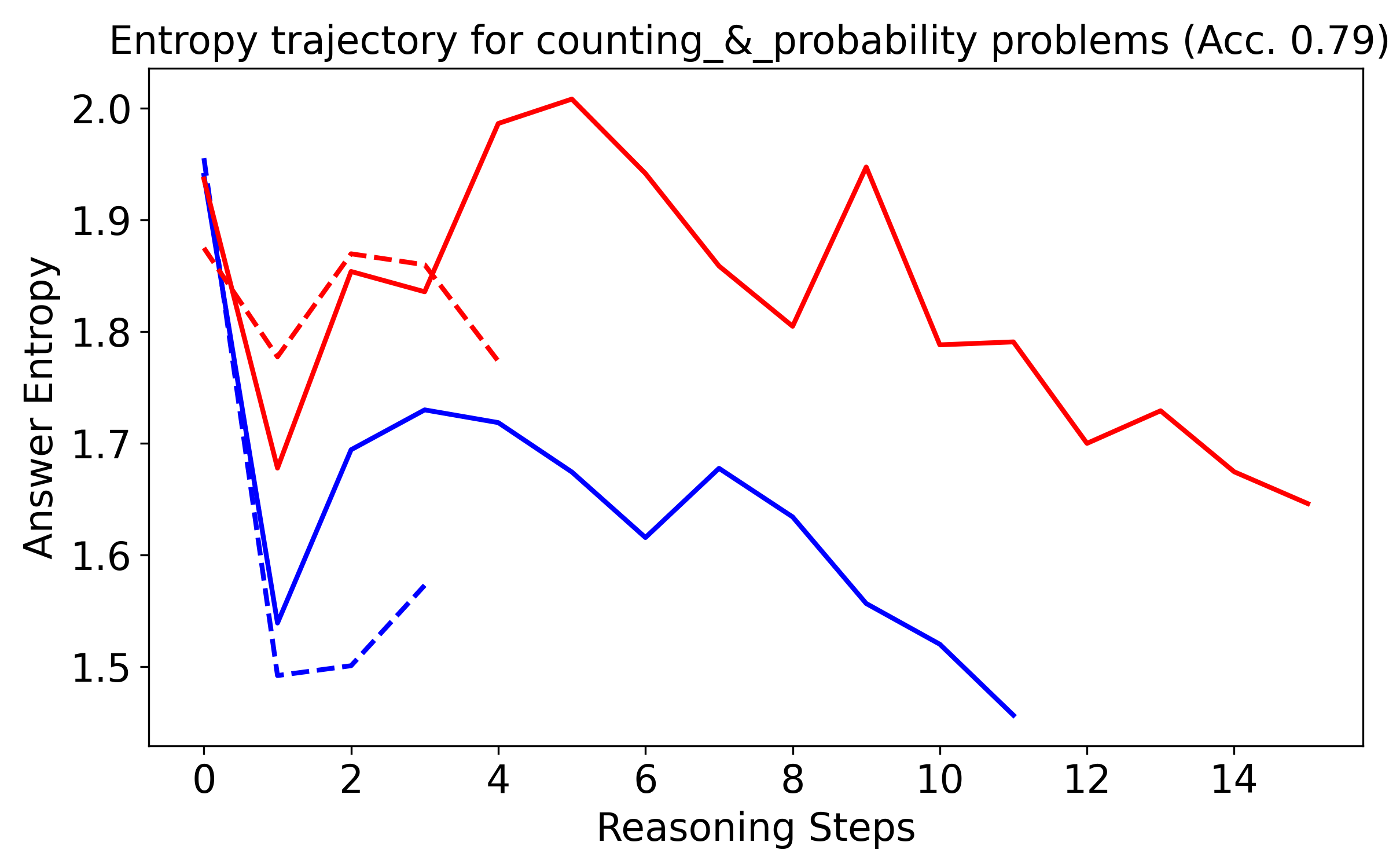}
    \caption{GPT-4o vs. Human on Counting \& Probability.}
  \end{subfigure}
  \hfill
  \begin{subfigure}[t]{0.47\textwidth}
    \centering
    \includegraphics[width=\textwidth]{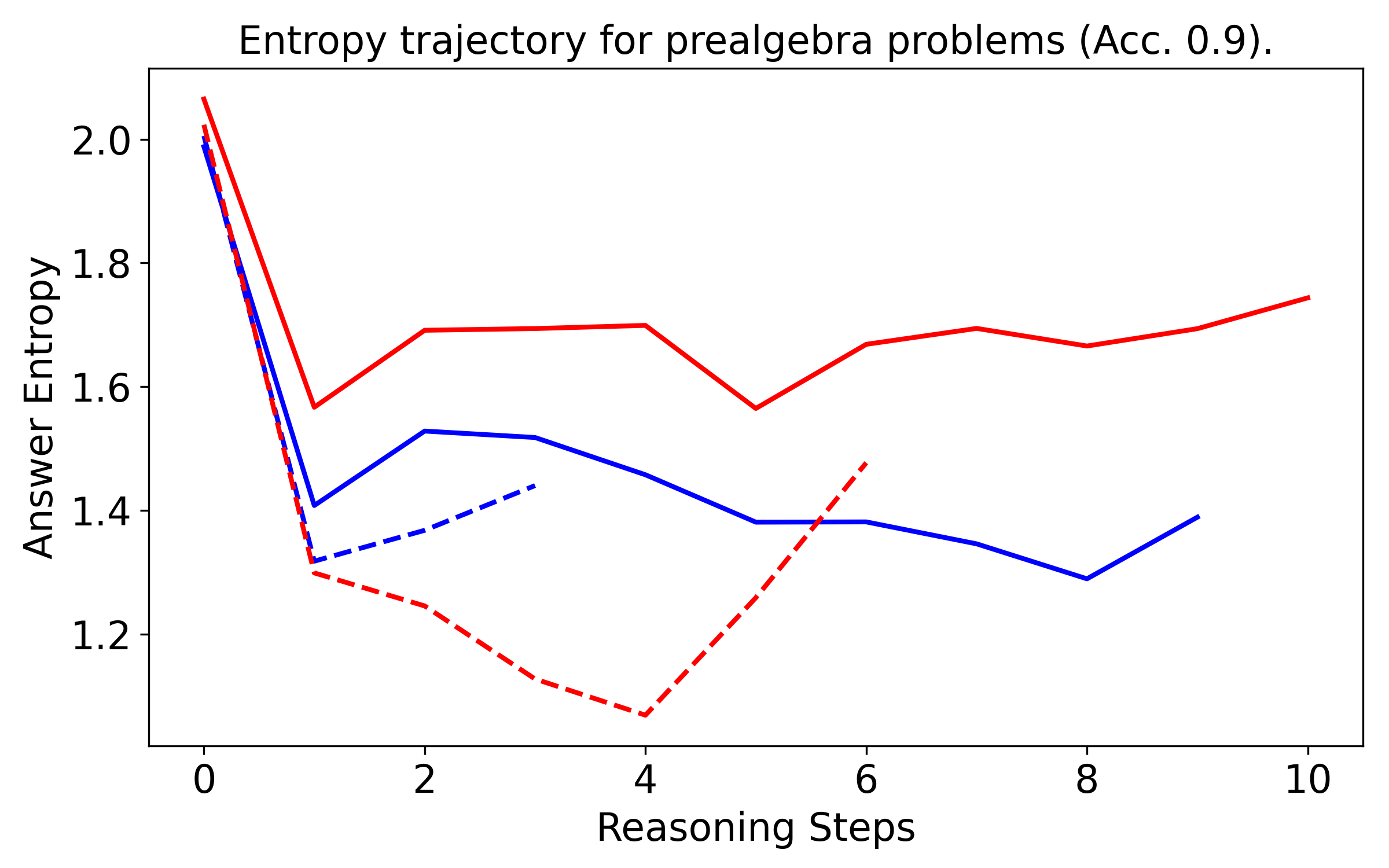}
    \caption{GPT-4o vs. Human Reasoning on Prealgebra.}
  \end{subfigure}
  \hfill
  \begin{subfigure}[t]{0.47\textwidth}
    \centering
    \includegraphics[width=\textwidth]{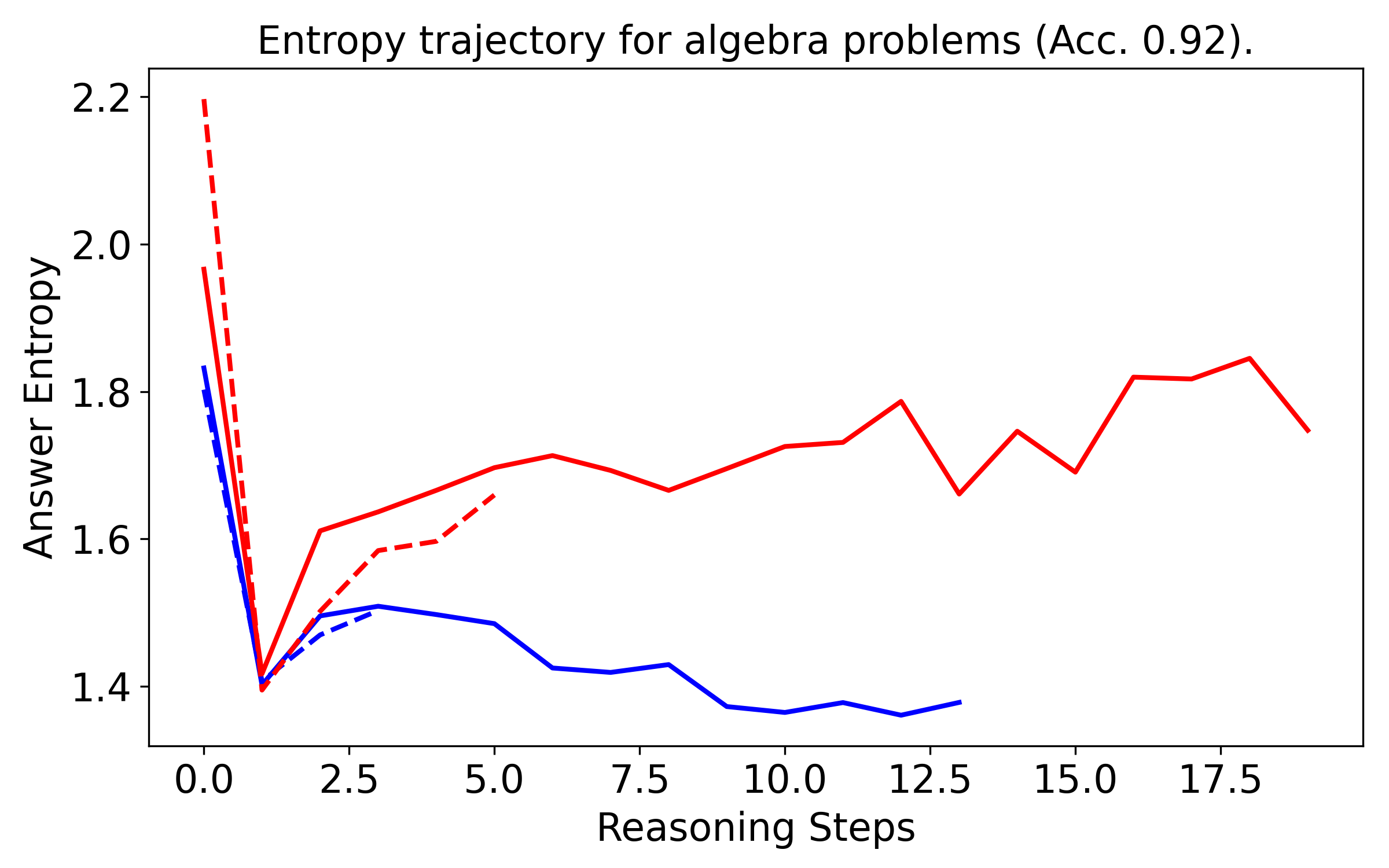}
    \caption{GPT-4o vs. Human Reasoning on Algebra.}
  \end{subfigure}
  \caption{Entropy variation across reasoning steps for GPT-4o. The overall trends are similar to Qwen2.5-32B: entropy is largely non-decreasing in low-accuracy categories but clearly decreases in high-accuracy ones. A key difference is that GPT-4o’s entropy drops rapidly and then plateaus earlier than Qwen2.5-32B (i.e., information gain diminishes earlier). GPT-4o also produces significantly longer reasoning chains than Qwen2.5-32B; this greater length may partly explain the prolonged plateau. }
  \label{fig:overall-gpt4o}
\end{figure*}

\end{document}